\PassOptionsToPackage{unicode}{hyperref}
\PassOptionsToPackage{hyphens}{url}
\documentclass[
  10pt,
]{article}
\usepackage{xcolor}
\usepackage[margin=1in]{geometry}
\usepackage{amsmath,amssymb}
\usepackage{iftex}
\ifPDFTeX
  \usepackage[T1]{fontenc}
  \usepackage[utf8]{inputenc}
  \usepackage{textcomp} 
\else 
  \usepackage{unicode-math} 
  \defaultfontfeatures{Scale=MatchLowercase}
  \defaultfontfeatures[\rmfamily]{Ligatures=TeX,Scale=1}
\fi
\usepackage{lmodern}
\ifPDFTeX\else
\fi
\IfFileExists{upquote.sty}{\usepackage{upquote}}{}
\IfFileExists{microtype.sty}{
  \usepackage[]{microtype}
  \UseMicrotypeSet[protrusion]{basicmath} 
}{}
\makeatletter
\@ifundefined{KOMAClassName}{
  \IfFileExists{parskip.sty}{%
    \usepackage{parskip}
  }{
    \setlength{\parindent}{0pt}
    \setlength{\parskip}{6pt plus 2pt minus 1pt}}
}{
  \KOMAoptions{parskip=half}}
\makeatother
\usepackage{color}
\usepackage{fancyvrb}

\DefineVerbatimEnvironment{Highlighting}{Verbatim}{commandchars=\\\{\}}
\newenvironment{Shaded}{}{}

\newcommand{\AttributeTok}[1]{\textcolor[rgb]{0.49,0.56,0.16}{#1}}

\newcommand{\BuiltInTok}[1]{\textcolor[rgb]{0.00,0.50,0.00}{#1}}

\newcommand{\ExtensionTok}[1]{#1}

\newcommand{\FunctionTok}[1]{\textcolor[rgb]{0.02,0.16,0.49}{#1}}

\newcommand{\NormalTok}[1]{#1}
\newcommand{\OperatorTok}[1]{\textcolor[rgb]{0.40,0.40,0.40}{#1}}

\newcommand{\VariableTok}[1]{\textcolor[rgb]{0.10,0.09,0.49}{#1}}

\usepackage{longtable,booktabs,array}
\usepackage{caption}
\usepackage{calc} 
\usepackage{etoolbox}
\makeatletter
\patchcmd\longtable{\par}{\if@noskipsec\mbox{}\fi\par}{}{}
\makeatother
\IfFileExists{footnotehyper.sty}{\usepackage{footnotehyper}}{\usepackage{footnote}}
\makesavenoteenv{longtable}
\usepackage{graphicx}
\makeatletter
\newsavebox\pandoc@box
\newcommand*\pandocbounded[1]{
  \sbox\pandoc@box{#1}%
  \Gscale@div\@tempa{\textheight}{\dimexpr\ht\pandoc@box+\dp\pandoc@box\relax}%
  \Gscale@div\@tempb{\linewidth}{\wd\pandoc@box}%
  \ifdim\@tempb\p@<\@tempa\p@\let\@tempa\@tempb\fi
  \ifdim\@tempa\p@<\p@\scalebox{\@tempa}{\usebox\pandoc@box}%
  \else\usebox{\pandoc@box}%
  \fi%
}
\def\fps@figure{htbp}
\makeatother
\providecommand{\tightlist}{%
  \setlength{\itemsep}{0pt}\setlength{\parskip}{0pt}}
\usepackage{longtable,booktabs,array}
\usepackage{ragged2e}
\usepackage{etoolbox}
\AtBeginEnvironment{longtable}{\footnotesize\RaggedRight}
\usepackage[htt]{hyphenat}
\renewcommand{\texttt}[1]{\begingroup\ttfamily\hyphenchar\font=`\-\relax\spaceskip=0pt\hskip0pt\allowbreak#1\endgroup}


\newenvironment{preprintabstract}{%
  \begin{list}{}{\setlength{\leftmargin}{9mm}\setlength{\rightmargin}{9mm}%
    \setlength{\labelwidth}{0pt}\setlength{\labelsep}{0pt}%
    \setlength{\itemindent}{0pt}\setlength{\listparindent}{0pt}%
    \setlength{\topsep}{3pt}\setlength{\parsep}{\parskip}}%
  \item\relax
}{\end{list}}
\usepackage{bookmark}
\IfFileExists{xurl.sty}{\usepackage{xurl}}{} 
\makeatletter
\@ifundefined{xmpquote}{}{}
\makeatother
\hypersetup{
  pdftitle={Finite-Horizon Fisher Memory in Two-Sided Power-Bounded Recurrent Systems},
  pdfauthor={Jeonghoon Lee, Attractor Dynamics Inc.},
  hidelinks,
  pdfcreator={LaTeX via pandoc}}

\title{Finite-Horizon Fisher Memory in Two-Sided Power-Bounded Recurrent
Systems}
\usepackage{etoolbox}
\makeatletter
\providecommand{\subtitle}[1]{
  \apptocmd{\@title}{\par {\large #1 \par}}{}{}
}
\makeatother
\subtitle{Normality, asymptotic information geometry, and downstream
storage}
\author{Jeonghoon Lee, Attractor Dynamics Inc.}
\date{30 September 2026}

\begin{document}
\maketitle

\subsection{Abstract}\label{abstract}

\begin{preprintabstract}

A noisy recurrent memory must select information during writing,
transfer it to a store, and preserve it after writing ends. We analyse
allocation, admission and retention in finite-horizon linear-Gaussian
systems. The directional Fisher memory \(M_n\) has a fixed trace budget,
\(\operatorname{tr}M_n=N\), at every horizon. Non-normality can
redistribute information across directions but cannot raise its
spherical average. A normal carrier satisfies \(M_n=I\) exactly. For
bi-power-bounded carriers, lag-wise information has uniform \(1/n\) tail
bounds. Using classical operator theory, we identify the limit of
\(M_n\) with the inverse of the Cesàro asymptotic limit of \(W^\top\),
obtain a commutant representation, and bound finite-horizon error using
the spectral-group separation.

Transfer through a time-varying coupling gives an end-to-end operator
\(M_{\mathrm{store}}\). An input direction that is optimal for the
writer need not be optimal for the store: selection depends on the store
objective. After writing ends, an invertible hold preserves the complete
stored Fisher matrix. Additive contamination bounded by \(\alpha\) times
the closure covariance retains at least \(1/(1+\alpha)\) of that matrix,
and a covariance-aware decoder attains the corresponding accuracy.

With the recurrent carriers held fixed, behavioural-loss training of
input masks and linear readouts approached this task-specific optimum in
160 runs, at a median normalized Rayleigh efficiency above \(0.998\)
against a random-direction baseline of \(0.14\) to \(0.26\). Empirical
binary accuracy matched the Gaussian prediction \(\Phi(a\sqrt J)\) to a
mean absolute error below \(0.002\) over more than four orders of
magnitude in \(J\). In a separate pre-specified study of 320 runs,
changing the designated input time changed the end-to-end operator. The
independently trained masks followed the corresponding objective in both
carrier types, in 16 of 16 draws.

Exact isolation preserved the modelled information across horizons while
continued coupling degraded it. A decoder fixed at its training horizon
fell to chance although that information was unchanged; inverse-adjoint
transport restored its sampled decisions to numerical precision. Both
studies used carriers that had been run at reduced budget during
development, so we report them as pre-specified validations rather than
blind holdouts. A third run of the same fixed design, on a block of
carriers not used before the run was committed, reproduced the
objective-specific result in 16 of 16 draws.

\end{preprintabstract}

\subsection{1. Introduction}\label{introduction}

A recurrent memory can retain its state and still give a poor answer.
Information may be concentrated in the wrong direction, fail to reach a
downstream store, or remain in the store while a fixed decoder loses
access to it. We separate these questions into allocation during
writing, admission to a store, and retention after writing ends.

The Fisher memory curve (FMC) of Ganguli, Huh and Sompolinsky {[}1{]}
describes the write stage for stable linear carriers with noise injected
inside the loop. For the input \(k\) steps ago, \(J(k)\) is the Fisher
information in the current state; its sum is a capacity. They show that
normal carriers have total capacity one and that any \(N\)-dimensional
carrier has capacity at most \(N\), with extensive capacity arising from
strongly non-normal feedforward constructions. Tiňo {[}12{]} and Kang,
Shirasaka and Suzuki {[}21{]} make the write direction explicit in
contracting reservoirs.

We use a finite-horizon formulation that remains defined at exact
isometry. The main analytic object is a matrix \(M_n\) whose Rayleigh
quotient is the total Fisher memory of a unit write direction. Its trace
is fixed, but its spectrum need not be. Noise placement matters:
receiver-noise communication capacity can benefit from non-normal
amplification {[}32{]}, whereas the in-loop Fisher trace here is fixed;
§10 compares the definitions.

This paper studies the finite-window analogue of the spatial Fisher
memory matrix of {[}1{]}, building on that stationary framework and on
Fisher-optimal input design {[}12, 21{]}. It makes three contributions:

\begin{itemize}
\tightlist
\item
  \textbf{Finite-horizon geometry.} We give an elementary horizon-two
  characterization of normality and identify the long-horizon Fisher
  operator of a two-sided power-bounded carrier with the inverse of the
  classical Cesàro asymptotic limit of its adjoint. From this
  identification, we obtain a commutant representation, uniform lag-wise
  consequences and explicit finite-horizon error control.
\item
  \textbf{Downstream-store objectives.} We separate write-stage
  direction selection from selection for a downstream store, whose
  coupling and closure covariance can change the leading direction.
\item
  \textbf{Task-specific learning and readout.} On fixed linear-Gaussian
  carriers, independently trained input masks follow the designated
  input-time objective under a fixed budget. A separate block of
  previously unused carriers reproduces this objective-specific
  separation. Storage interventions and sampled decoder tests
  distinguish retained information from decoder access.
\end{itemize}

We connect Fisher allocation, downstream-store selection and post-write
retention through a quantitative finite-horizon analysis and behavioural
validation. The operator-theoretic ingredients and
information-preservation identities are established results, credited in
the \emph{Classical ingredients} subsection before §3.3. Our
contribution is their specialization to Fisher memory, the
finite-horizon error control, and the validation across these stages.
Deterministic losslessness can coexist with directional concentration,
but continuing loop noise rules out a non-vanishing oldest-lag Fisher
floor.

We derive a finite-horizon error bound controlled by the spectral-group
gap and require a direct finite-\(n\) cross-check. Near-degenerate
eigenvalue groups can delay the Cesàro projection, so the asymptotic
formula alone does not certify an operating horizon.

The post-write claim also depends on the channel and decoder. When no
further coupling or noise enters the store, a known invertible hold,
including a contraction, changes coordinates but preserves exact
covariance-aware Fisher information. Bounded additive contamination
gives a quantitative retained-information guarantee. Neither result
makes a fixed, quantized or covariance-mismatched decoder invariant. We
therefore also check sampled decisions.

The claims apply to this linear-Gaussian class. Non-normal write
dynamics can concentrate a trace-constrained Fisher budget into selected
directions; an end-to-end operator can improve the direction admitted to
a finite store; and the specified post-write channel determines how that
geometry is preserved or degraded. Whether trained nonlinear models
discover these directions remains untested. Useful task accuracy also
requires sufficient absolute signal-to-noise ratio. Applicability to
other memory architectures must be assessed for their own channels and
objectives.

\subsection{2. Finite-horizon Fisher-memory
setting}\label{finite-horizon-fisher-memory-setting}

Carrier \(x_{t+1}=Wx_t+v\,s_t+z_t\) with scalar input \(s_t\) (the
quantity whose past values the state carries information about),
\(z_t\sim\mathcal N(0,\epsilon I)\), \(\epsilon=1\), unit write vector
\(v\), \(N=32\). With \(C_n=\sum_{j=0}^{n-1}W^j(W^j)^\top\succeq I\),
\[J_n(k)=v^\top(W^\top)^kC_n^{-1}W^kv,\qquad J_{tot,n}(v)=v^\top M_nv,\qquad M_n=\sum_{k=0}^{n-1}(W^\top)^kC_n^{-1}W^k.\]

The spatial Fisher memory operator, the result that a normal carrier
spreads a unit of directional capacity isotropically while a non-normal
one concentrates it, and the selection of a write direction as a leading
eigenvector are due to Ganguli, Huh and Sompolinsky {[}1{]} in the
stationary setting, and Kang, Shirasaka and Suzuki {[}21{]} use the
leading-eigenvector rule for input-mask design. What is finite-horizon
here is the truncation of \(C_n\) and \(M_n\) at \(n\), which keeps the
objects defined at exact isometry, where no stationary curve exists.

Throughout, the initial state is known, \(x_0=0\); an unknown \(x_0\)
with covariance \(\Sigma_0\) would add \(W^n\Sigma_0(W^n)^\top\) to
\(C_n\). The noise variance is \(\epsilon=1\); for general \(\epsilon\)
every Fisher quantity scales by \(1/\epsilon\): the Fisher matrix is
\(M_n/\epsilon\), of trace \(N/\epsilon\), while \(M_n\) and
\(\mathrm{tr}M_n=N\) do not depend on \(\epsilon\). \(J_n(k)\) is the
diagonal element of the Fisher information matrix in the input sequence,
with all other inputs treated as known, which is the convention of
{[}1{]}. This is the finite-horizon object; for \(\rho(W)=1\) there is
no stationary FMC, and nothing below uses one. We report totals at
\(n\in\{32,128,512,2048\}\) and the \emph{oldest-lag finite-horizon
Fisher information} \(J_n(n-1)\) at \(n\le4096\), together with
\(nJ_n(n-1)\).

\textbf{Definition (bi-power-bounded).} An invertible \(W\) is
bi-power-bounded if \(K_+=\sup_{j\ge0}\|W^j\|_2<\infty\) and
\(K_-=\sup_{j\ge0}\|W^{-j}\|_2<\infty\). \(SQS^{-1}\) with \(Q\)
orthogonal (\(K_\pm\le\mathrm{cond}(S)\)) and \(e^{JH}\) with
\(H\succ0\) (similar to orthogonal through \(H^{1/2}\)) are
bi-power-bounded; \(0.9Q\) is invertible but not bi-power-bounded.

\textbf{Certificates.} Normality defect \(\|WW^\top-W^\top W\|_F\);
symplectic defect \(\|W^\top JW-J\|_F\). We report two finite-window
gains separately: the forward gain
\(\sigma^{+}_{n}=\max_{0\le k\le n}\|W^{k}\|_2\), a lower bound on
\(K_+\) over the measured horizon, and the inverse gain
\(\sigma^{-}_{n}=\max_{0\le k\le n}\|W^{-k}\|_2\), a lower bound on
\(K_-\). Main-text plots use \(\sigma^{+}_{4096}\); Appendix B tabulates
both and labels each column accordingly.

\subsection{3. Exact directional
geometry}\label{exact-directional-geometry}

\subsubsection{3.1 Trace budget}\label{trace-budget}

\[
\operatorname{tr}M_n
=\operatorname{tr}\!\left(C_n^{-1}\sum_{k<n}W^k(W^k)^\top\right)
=\operatorname{tr}(C_n^{-1}C_n)=N .
\]

Hence the spherical average of \(v^\top M_nv\) over unit write
directions is one, and its range is the interval
\([\lambda_{\min}(M_n),\lambda_{\max}(M_n)]\). Non-normality can
redistribute the budget but cannot raise its direction average. At
\(n=1\), \(M_1=I\) for every carrier.

This is the finite-horizon form of the spatial identity
\(\operatorname{tr}J_s=N\) stated by Ganguli, Huh and Sompolinsky
{[}1{]}. Let \(H=[I,W,\ldots,W^{n-1}]\), so \(C_n=HH^\top\). Then
\(\Pi=H^\top C_n^{-1}H\) is a rank-\(N\) orthogonal projector whose
diagonal blocks sum to \(M_n\). For a unit direction \(v\),
\(J_n(k;v)=(e_k\otimes v)^\top\Pi(e_k\otimes v)\), where \(e_k\) selects
lag block \(k\). This is a directional leverage; after choosing \(v\) as
a coordinate axis, it is a diagonal hat-matrix leverage in the sense of
Hoaglin and Welsch {[}33{]}.

\subsubsection{3.2 Normal isotropy at finite
horizon}\label{normal-isotropy-at-finite-horizon}

If \(W\) is normal, \(C_n\) and every \((W^\top)^kC_n^{-1}W^k\) are
simultaneously diagonalizable. Their eigenvalues at a carrier eigenvalue
\(\mu\) are \[
\frac{|\mu|^{2k}}{\sum_{j<n}|\mu|^{2j}},
\] which sum to one over \(k<n\). Therefore \[
\boxed{M_n=I}
\] for every finite horizon and every spectral radius. This
finite-horizon identity needs no stability assumption; the original
stationary theorem {[}1{]} does.

\textbf{Proposition 3.2b (horizon-two characterization of normality).}
At \(n=2\), \[
M_2=I+(I+WW^\top)^{-1}-(I+W^\top W)^{-1},
\] so \(M_2=I\) if and only if \(W\) is normal. Since
\(\operatorname{tr}M_2=N\), every non-normal \(W\) has at least one
direction with \(J_{tot,2}>1\) and one with \(J_{tot,2}<1\).

\emph{Proof.} \(C_2=I+WW^\top\), and the push-through identity gives the
displayed form; the two inverses coincide exactly when
\(WW^\top=W^\top W\). The trace budget of §3.1 then forces directions on
both sides of one. Appendix A.7 gives the steps. \(\square\)

Proposition 3.2b is an elementary converse, and we claim no priority for
it. We include it because it is the simplest finite-horizon form of the
normal/non-normal dichotomy. It is only an existence statement: it says
nothing about the size of the advantage at an operating horizon, or
about whether a finite-budget optimizer can reach it.

\subsubsection{Classical ingredients, and what is being
identified}\label{classical-ingredients-and-what-is-being-identified}

The limit theorems below rest on two known facts, and the limit object
is a known matrix.

\emph{Two-sided power boundedness.} A finite-dimensional matrix whose
positive and negative powers are uniformly bounded is similar to an
orthogonal matrix, \(W=SQS^{-1}\) with \(Q^\top Q=I\). This is classical
operator theory in the Sz.-Nagy similarity tradition; Gehér {[}30, 31{]}
gives the matrix statement we use.

\emph{Cesàro asymptotic limits.} For a power-bounded matrix the Cesàro
averages of \((T^\top)^jT^j\) converge in norm, and for matrices similar
to unitaries the positive limit is characterized together with its
inverse-eigenvalue trace constraint. These are Theorems 1 and 2 in the
arXiv version of {[}30{]} and Theorems 3.1 and 3.2 in {[}31, Ch. 3{]};
the mean-ergodic theorem alone would not give the matrix statement we
need.

\emph{The identification.} Write \(G_n=C_n/n\) and let
\(\mathcal A_C(T)=\lim_n n^{-1}\sum_{j<n}(T^\top)^jT^j\) be the
classical Cesàro limit. Then \(G=\lim_nG_n=\mathcal A_C(W^\top)\succ0\),
and boundedness of the powers gives
\(WG_nW^\top-G_n=(W^n(W^n)^\top-I)/n\to0\), hence the fixed-point
relations \[
WGW^\top=G,\qquad W^\top G^{-1}W=G^{-1}.
\] Since \(C_n^{-1}=n^{-1}G_n^{-1}\), the finite Fisher operator obeys
\(M_n-G^{-1}=n^{-1}\sum_{k<n}(W^\top)^k(G_n^{-1}-G^{-1})W^k\), so with
\(K_+=\sup_{k\ge0}\|W^k\|_2\) \[
\boxed{\bigl\lVert M_n-G^{-1}\bigr\rVert_2\le K_+^2\bigl\lVert G_n^{-1}-G^{-1}\bigr\rVert_2\longrightarrow0,}
\] and the same estimate bounds
\(\sup_{0\le k<n}\lvert nJ_n(k;v)-v^\top G^{-1}v\rvert\) uniformly in
\(v\). The central identification is therefore \[
\boxed{M_\infty=\mathcal A_C(W^\top)^{-1},}
\] which for \(W=SQS^{-1}\) is exactly \(G=S\bar AS^\top\) and
\(M_\infty=S^{-\top}\bar A^{-1}S^{-1}\), the form stated in Theorem
3.3b. Thus \(M_\infty\) is the inverse of a known asymptotic object, and
what follows is its Fisher-memory interpretation together with the
finite-horizon control of §3.4. A numerical check of this identification
on the carriers used here is included in the reproduction code.

Thus \(\operatorname{tr}M_\infty=N\) is precisely Gehér's
inverse-eigenvalue trace constraint applied to \(\mathcal A_C(W^\top)\)
{[}30{]}. More precisely, \(M_\infty=UU^\top\) for a real square matrix
\(U\) with unit columns: it is the sum of their outer products, while
their Gram matrix is \(U^\top U\), orthogonally similar to \(M_\infty\).
Every real positive-definite matrix of trace \(N\) is attainable as
\(M_\infty\) for a real bi-power-bounded carrier; Appendix A.4 gives a
construction.

\subsubsection{3.3 Uniform Fisher-tail bounds and the Cesàro--commutant
limit}\label{uniform-fisher-tail-bounds-and-the-cesuxe0rocommutant-limit}

\textbf{Theorem 3.3a (uniform tail bounds).} If \(W\) is
bi-power-bounded, with \[
K_+=\sup_{j\ge0}\|W^j\|_2,\qquad K_-=\sup_{j\ge0}\|W^{-j}\|_2,
\] then for every unit \(v\), \(n\ge1\), and \(0\le k<n\), \[
\boxed{\frac{1}{nK_+^2K_-^2}\le J_n(k)\le\frac{K_+^2K_-^2}{n}.}
\] Thus \(J_n(k)=\Theta(n^{-1})\) uniformly over lags and directions
under continuing isotropic loop noise.

\textbf{Theorem 3.3b (Cesàro--commutant limit).} Every
finite-dimensional bi-power-bounded real matrix can be written \[
W=SQS^{-1},\qquad Q^\top Q=I.
\] Set \[
A=S^{-1}S^{-\top},\qquad
A_n=\frac1n\sum_{j=0}^{n-1}Q^jAQ^{-j},\qquad
\bar A=\Pi_{\operatorname{Comm}(Q)}(A).
\] Then \[
\boxed{M_n\longrightarrow M_\infty=S^{-\top}\bar A^{-1}S^{-1}}
\] and, uniformly over finite-horizon lags, \[
\boxed{\sup_{0\le k<n}\left|nJ_n(k;v)-v^\top M_\infty v\right|\longrightarrow0.}
\] At large horizons, every lag receives approximately the same \(1/n\)
share of the directional total \(v^\top M_\infty v\).

\subsubsection{3.4 Finite-horizon
certification}\label{finite-horizon-certification}

The Cesàro estimate below is the standard spectral-gap rate for unitary
means, applied to the conjugation map \(A\mapsto QAQ^\top\) on matrices
with the Frobenius inner product. Kachurovskii surveys this spectral
approach {[}34{]}, and Aloisio et al.~state the discrete-time gap bound
explicitly {[}35, Theorem 2{]}; Short and Farrelly give the related
continuous-time gap estimate {[}36, Eqs. 11--12{]}.

Let the distinct complex eigenvalue groups of \(Q\) have projectors
\(P_\lambda\) and define \[
\Delta=\min_{\lambda\ne\mu}|1-\lambda\bar\mu|.
\] For \[
\bar A=\sum_\lambda P_\lambda A P_\lambda,
\] the finite Cesàro average obeys \[
\boxed{
\|A_n-\bar A\|_F
\le
\delta_n
:=
\frac{2}{n\Delta}\|A-\bar A\|_F .
}
\] If \[
\eta_n:=\|\bar A^{-1}\|_2\,\delta_n<1,
\] then \[
\boxed{
\|M_n-M_\infty\|_2
\le
\|S^{-1}\|_2^2
\frac{\|\bar A^{-1}\|_2^2\delta_n}{1-\eta_n}.
}
\] The bound is sufficient and can be conservative. It makes the
dependence on near-degenerate eigenvalue groups explicit and provides a
rejection rule: when \(n\Delta\) is too small, a finite Cesàro
construction is not certified as the commutant projection.

Across the eight paired \(c=10\) instances used below, the actual
relative Frobenius error \(\|M_{2048}-M_\infty\|_F/\|M_\infty\|_F\) had
median \(8.44\times10^{-6}\) and range \(3.56\times10^{-6}\) to
\(1.79\times10^{-5}\); every actual operator-norm error lay below the
reported sufficient bound. A four-dimensional near-degenerate control
with eigen-angle separation \(10^{-4}\) at \(n=1000\) had \[
\|A_n-\bar A\|_F=1.599,\qquad
\|A_{2n}-A_n\|_F=0.0799,\qquad
\|[A_n,Q]\|_F=0.00102,
\] while \(2/(n\Delta)\approx20\). Small two-scale and commutator
residuals alone would therefore be an unsafe certificate.

\subsubsection{3.5 Two-dimensional elliptic
corollary}\label{two-dimensional-elliptic-corollary}

For \[
W=D^{-1}R(\theta)D,\qquad D=\operatorname{diag}(1,\sqrt c),\qquad \theta\notin\pi\mathbb Z,
\] the symmetric commutant projection is scalar, giving \[
\boxed{
M_\infty=\frac{2}{1+c}\operatorname{diag}(1,c),\quad
\lambda_{\max}=\frac{2c}{1+c},\quad
\lambda_{\min}=\frac{2}{1+c}.
}
\] At \(\theta\in\pi\mathbb Z\), \(W=\pm I\) and \(M_n=I\). The ceiling
two is the block trace budget, not a generic consequence of
symplecticity.

\includegraphics[width=1\linewidth,height=\textheight,keepaspectratio]{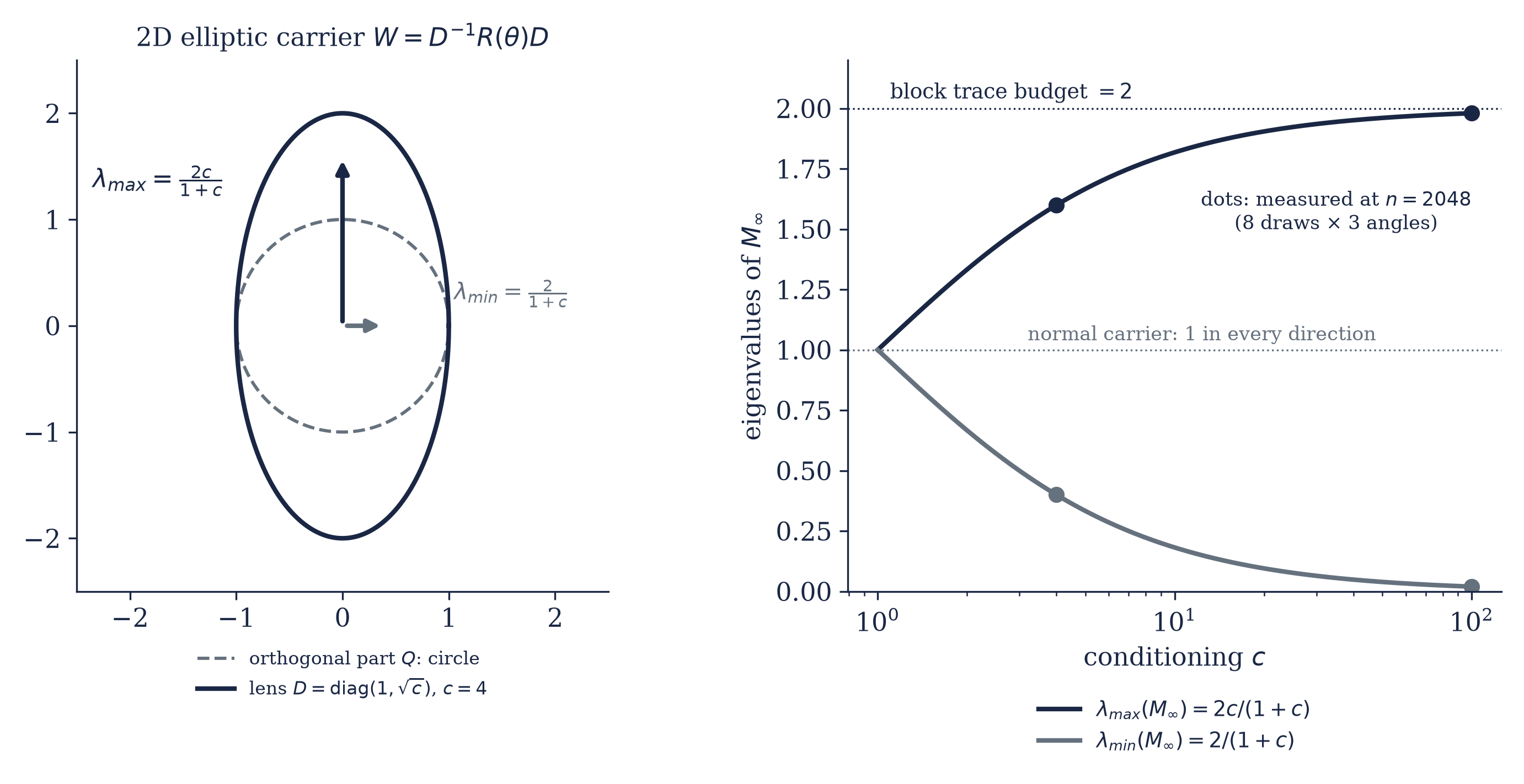}

\emph{Figure 1. The two-dimensional elliptic family and its limiting
directional spectrum.}

\subsection{4. Replicated directional-capacity
census}\label{replicated-directional-capacity-census}

\(M_n\) is evaluated at \(n=2048\), and the oracle direction is fixed at
2048. Values are medians with {[}min, max{]} over draws, and full
quartiles are in the released CSV. Each draw randomizes the orthogonal
\(Q\), the left and right singular bases of \(S\) (its singular values
are fixed at \(\mathrm{geomspace}(1,c,N)\)), the orthogonal mixer of the
2D family, and the eigenbasis of \(H\) and the orthogonal-symplectic
mixer of the Hamiltonian family. The write vector is not randomized,
because the oracle direction is computed from \(M_{2048}\).

{\def\LTcaptype{none} 
\begin{longtable}[]{@{}
  >{\raggedright\arraybackslash}p{(\linewidth - 10\tabcolsep) * \real{0.1667}}
  >{\raggedright\arraybackslash}p{(\linewidth - 10\tabcolsep) * \real{0.1667}}
  >{\raggedright\arraybackslash}p{(\linewidth - 10\tabcolsep) * \real{0.1667}}
  >{\raggedright\arraybackslash}p{(\linewidth - 10\tabcolsep) * \real{0.1667}}
  >{\raggedright\arraybackslash}p{(\linewidth - 10\tabcolsep) * \real{0.1667}}
  >{\raggedright\arraybackslash}p{(\linewidth - 10\tabcolsep) * \real{0.1667}}@{}}
\toprule\noalign{}
\begin{minipage}[b]{\linewidth}\raggedright
Carrier
\end{minipage} & \begin{minipage}[b]{\linewidth}\raggedright
cond
\end{minipage} & \begin{minipage}[b]{\linewidth}\raggedright
\(\sigma_{4096}=\max_{j\le4096}\|W^j\|_2\), med
\end{minipage} & \begin{minipage}[b]{\linewidth}\raggedright
\(\lambda_{max}\) (oracle) med {[}min, max{]}
\end{minipage} & \begin{minipage}[b]{\linewidth}\raggedright
\(\lambda_{min}\) med
\end{minipage} & \begin{minipage}[b]{\linewidth}\raggedright
\(nJ_n(n-1)\) at \(n=4096\), med
\end{minipage} \\
\midrule\noalign{}
\endhead
\bottomrule\noalign{}
\endlastfoot
Haar orthogonal & 1 & 1.00 & 1.000 {[}1.000, 1.000{]} & 1.000 & 1.000 \\
\(SQS^{-1}\) & 2 & 1.82 & 1.831 {[}1.774, 1.852{]} & 0.470 & 1.831 \\
& 5 & 4.14 & 3.139 {[}3.040, 3.298{]} & 0.133 & 3.138 \\
& 10 & 7.88 & 4.312 {[}4.098, 4.639{]} & 0.045 & 4.310 \\
& 20 & 15.0 & 5.491 {[}5.246, 5.637{]} & 0.014 & 5.488 \\
& 35 (held-out) & 25.6 & 6.107 {[}5.892, 6.331{]} & 0.0054 & 6.105 \\
& 50 & 35.1 & 6.563 {[}6.343, 6.880{]} & 0.0029 & 6.559 \\
& 100 & 68.9 & 7.720 {[}7.366, 8.061{]} & 0.0008 & 7.713 \\
2D elliptic, 4 angles & 4 & 2.00 & 1.600 {[}1.600, 1.600{]} & 0.400 &
1.600 \\
2D elliptic, 3 angles & 100 & 10.0 & 1.980 {[}1.980, 1.980{]} & 0.0198 &
1.980 \\
\(e^{JH}\), 8×8 blocks & 4 & 1.70 & 1.595 {[}1.468, 1.714{]} & 0.549 &
1.595 \\
& 100 & 4.94 & 2.771 {[}2.147, 3.351{]} & 0.114 & 2.771 \\
\end{longtable}
}

\includegraphics[width=0.95\linewidth,height=\textheight,keepaspectratio]{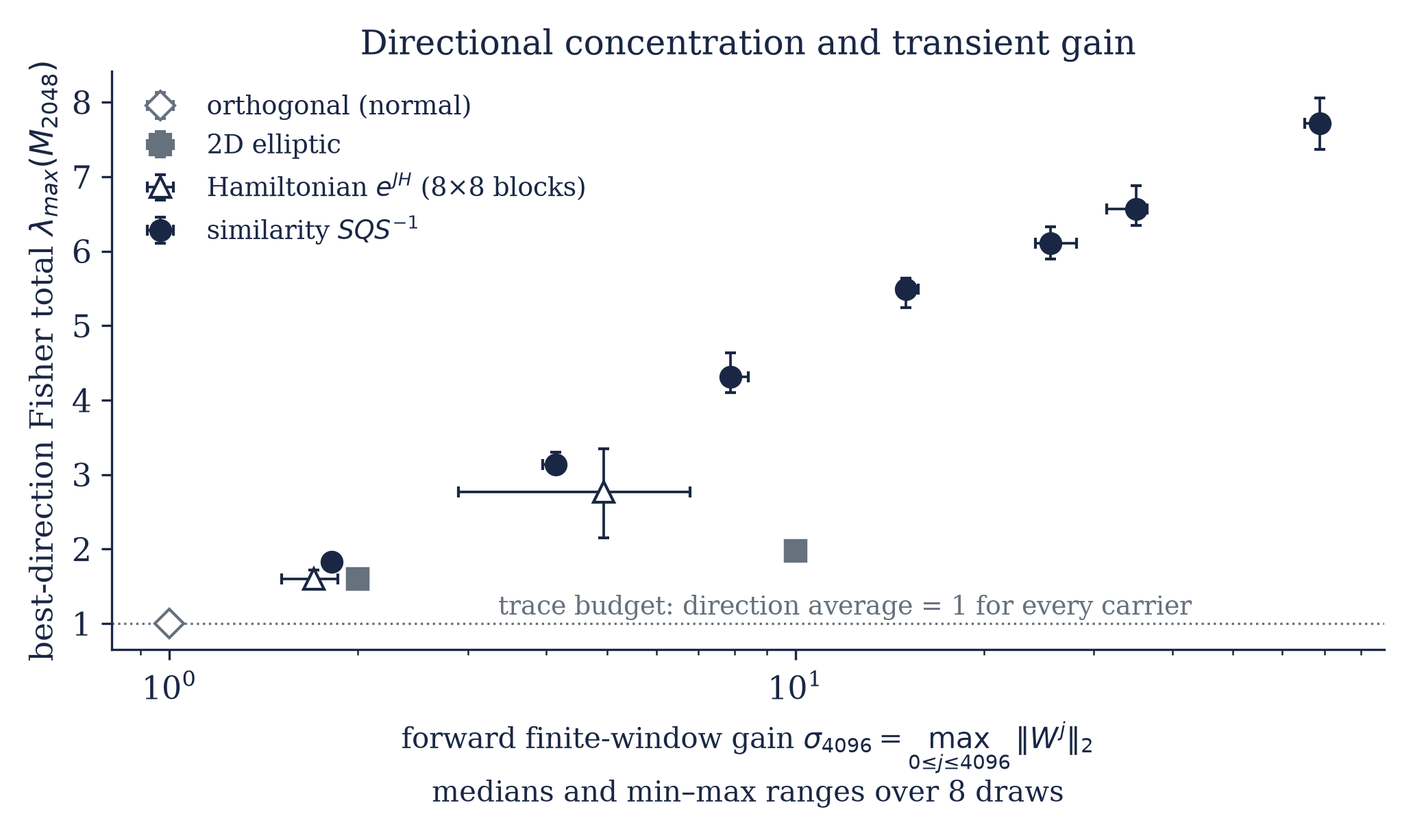}

\emph{Figure 2. Replicated census: best-direction Fisher total
\(\lambda_{max}(M_{2048})\) against the measured forward finite-window
gain \(\sigma_{4096}=\max_{0\le j\le4096}\|W^j\|_2\), with medians and
min--max ranges over eight draws per configuration. The direction
average is one for every point.}

\textbf{Interpretation.} (a) Every instance obeys the finite-window
bounds of §3.3 at every lag, and \(nJ_n(n-1)\) equals \(\lambda_{max}\)
within 2\% at \(n\ge512\) on all 128 instances, consistent with the
asymptotic theorem. The theorem alone does not specify that tolerance at
that horizon; the finite-horizon certificate depends on the
spectral-group separation and the conditioning.

\begin{enumerate}
\def\labelenumi{(\alph{enumi})}
\setcounter{enumi}{1}
\item
  Normality rules out directional concentration above one (§3.2), and by
  Proposition 3.2b every non-normal carrier has some. That is an
  existence statement at horizon two. It does not guarantee a large,
  robust or learnable advantage at a chosen operating horizon: this
  section measures the size, and §9 discusses learnability.
  Non-normality alone does not make the gain large: 8×8 Hamiltonian
  blocks at \(c=4\) reach only 1.6.
\item
  In the \(SQS^{-1}\) family the oracle capacity is monotone in
  conditioning; an exploratory fit of configuration medians,
  \(\lambda_{max}\approx0.91+1.62\ln\sigma_{4096}\) (train RMSE 0.10 on
  six conditionings), predicted the held-out \(c=35\) median to within
  0.05 (6.156 predicted, 6.107 observed). This is one family, and the
  fit is exploratory.
\end{enumerate}

\subsubsection{\texorpdfstring{4.1 The positive-definite Hamiltonian
exponential limits gain by
\(\sqrt{\kappa(H)}\)}{4.1 The positive-definite Hamiltonian exponential limits gain by \textbackslash sqrt\{\textbackslash kappa(H)\}}}\label{the-positive-definite-hamiltonian-exponential-limits-gain-by-sqrtkappah}

\(e^{JH}=H^{-1/2}e^{K}H^{1/2}\) with \(K\) skew-symmetric, so the
similarity transform is \(H^{1/2}\) and \(K_\pm\le\sqrt{\kappa(H)}\);
the observed \(\sigma_{4096}\) stays below \(\sqrt{\kappa(H)}\) (medians
1.70 at \(c=4\), 4.94 at \(c=100\), against 2 and 10), which is this
parametrization, not a property of symplecticity. Block trace budgets
bound \(\lambda_{max}\) by the block size (replicated: 1.98 of 2 and
2.77 of 8 at \(c=100\); the 4×4 value 1.91 of 4 is from the
single-instance run of 2026-09-02 and is not replicated). At matched
\(\sigma_{4096}\) the Hamiltonian and similarity families are close
(\(\sigma\approx4.1\): 3.14 versus \(\sigma\approx4.9\): 2.77). A
symplectic similarity family \(SQS^{-1}\) with \(S\in Sp\) would be
governed by \(\mathrm{cond}(S)\) and is not measured here.

\subsection{5. From write geometry to store
geometry}\label{from-write-geometry-to-store-geometry}

\subsubsection{5.1 Time-varying write--store
model}\label{time-varying-writestore-model}

Let the state be \(x=(x_w,x_s)\), where the write block has dimension
\(N\) and the store has dimension \(d\). During a finite write interval,
\[
x_{w,t+1}=W_wx_{w,t}+v s_t+z_t,
\] \[
x_{s,t+1}=K_tx_{w,t}+\Phi_tx_{s,t}.
\] For an input written at step \(t\), let
\(L_t:\mathbb R^N\to\mathbb R^d\) be its accumulated transfer into the
store at closure, and let \(C_{ss}\) be the noise-normalized closure
covariance of the store.

For vector inputs \(u_t\) and an endpoint
\(y\sim\mathcal N\bigl(\sum_tL_tu_t,\;C_{ss}\bigr)\), the joint Fisher
information in the input \emph{sequence} has time blocks \[
\mathcal I_{tu}=L_t^\top C_{ss}^{-1}L_u ,
\] and for scalar inputs written along a single direction \(v\) the
corresponding temporal matrix has entries
\(v^\top L_t^\top C_{ss}^{-1}L_uv\). The operator below sums the
diagonal time blocks into a spatial design operator on write directions.
It is therefore not the complete joint temporal Fisher matrix. Nor is it
an observability or constructibility Gramian: relating it to a Gramian
in the sense of {[}29{]} first requires mapping the input space, the
observation and the time aggregation.

Define the end-to-end store operator \[
\boxed{
M_{\mathrm{store}}=\sum_{t\in I}L_t^\top C_{ss}^{-1}L_t.
}
\] For a unit write direction \(v\), \(v^\top M_{\mathrm{store}}v\) is
the total conditioned Fisher information admitted to the store over the
input set \(I\). It depends on the write dynamics, coupling, write
interval, store map during the interval, and closure covariance.

\textbf{Proposition 5.1 (store budget).} If the closure covariance can
be written \[
C_{ss}=\sum_{r\in\mathcal N}L_rL_r^\top+C_{\mathrm{add}},\qquad C_{\mathrm{add}}\succeq0,
\] and \(I\subseteq\mathcal N\), then \[
\boxed{\operatorname{tr}M_{\mathrm{store}}\le d.}
\] Equality holds in the full-rank pure inherited-noise case when the
signal and noise index sets induce the same covariance. The write-block
identity \(\operatorname{tr}M_n=N\) does not automatically transfer to
the store.

\subsubsection{5.2 Paired write-path × isolation
factorial}\label{paired-write-path-isolation-factorial}

The paired experiment uses \(N=32\), \(d=16\), a 24-step write window,
the same Haar orthogonal factor \(Q\) within each pair, and either the
normal writer \(Q\) or the conditioned writer \(SQS^{-1}\) with
\(\kappa(S)=10\). The store coupling is either left open or set to zero
after the write window. No process noise enters the store directly.

\includegraphics[width=1\linewidth,height=\textheight,keepaspectratio]{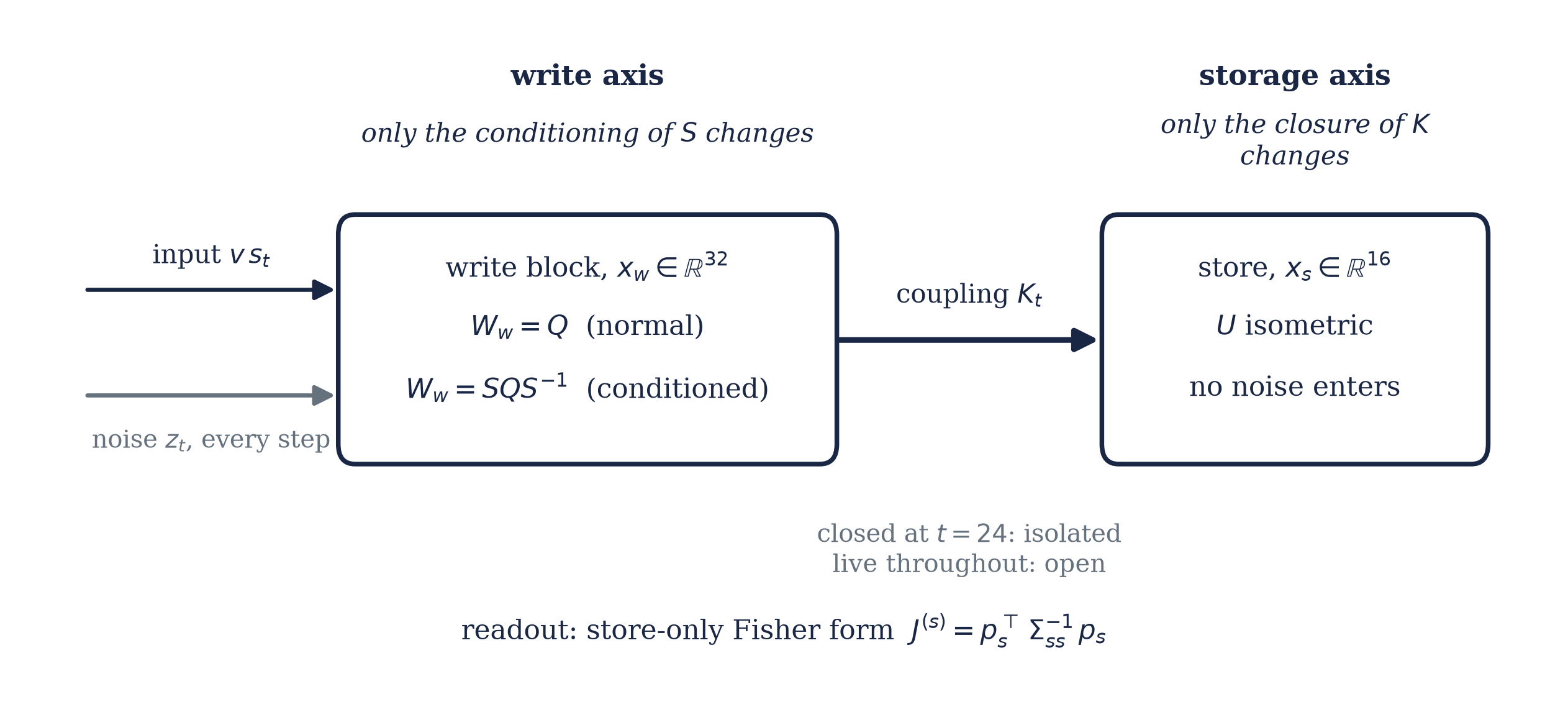}

\emph{Figure 3. The paired factorial. The paired cells share the same
Haar orthogonal factor \(Q\); the conditioned writer applies a sampled
similarity transform \(S\) with \(\kappa(S)=10\), which is not a scalar
change of condition number alone. The storage intervention changes only
whether the coupling into the store closes after the write window.}

At \(n=4096\), store-only medians over eight paired draws were:

{\def\LTcaptype{none} 
\begin{longtable}[]{@{}
  >{\raggedright\arraybackslash}p{(\linewidth - 8\tabcolsep) * \real{0.1667}}
  >{\raggedright\arraybackslash}p{(\linewidth - 8\tabcolsep) * \real{0.1667}}
  >{\raggedleft\arraybackslash}p{(\linewidth - 8\tabcolsep) * \real{0.2222}}
  >{\raggedleft\arraybackslash}p{(\linewidth - 8\tabcolsep) * \real{0.2222}}
  >{\raggedleft\arraybackslash}p{(\linewidth - 8\tabcolsep) * \real{0.2222}}@{}}
\toprule\noalign{}
\begin{minipage}[b]{\linewidth}\raggedright
write carrier
\end{minipage} & \begin{minipage}[b]{\linewidth}\raggedright
storage
\end{minipage} & \begin{minipage}[b]{\linewidth}\raggedleft
direction
\end{minipage} & \begin{minipage}[b]{\linewidth}\raggedleft
\(J^{(s)}_{\mathrm{tot}}\)
\end{minipage} & \begin{minipage}[b]{\linewidth}\raggedleft
oldest stored input
\end{minipage} \\
\midrule\noalign{}
\endhead
\bottomrule\noalign{}
\endlastfoot
normal \(Q\) & open & conditioned-writer oracle & 0.405 & 0.00008 \\
normal \(Q\) & isolated & conditioned-writer oracle & 0.421 & 0.03375 \\
conditioned \(SQS^{-1}\) & open & conditioned-writer oracle & 2.290 &
0.00056 \\
conditioned \(SQS^{-1}\) & isolated & conditioned-writer oracle & 2.509
& 0.18823 \\
\end{longtable}
}

The oracle is selected from the conditioned write-block matrix
\(M_{2048}\) and shared with the paired normal cell. It is a matched
directional intervention, not either store cell's independently
optimized direction. In the random direction the ordering reverses
(isolated medians \(0.597\) normal and \(0.439\) conditioned),
consistent with the fixed trace budget.

\includegraphics[width=0.95\linewidth,height=\textheight,keepaspectratio]{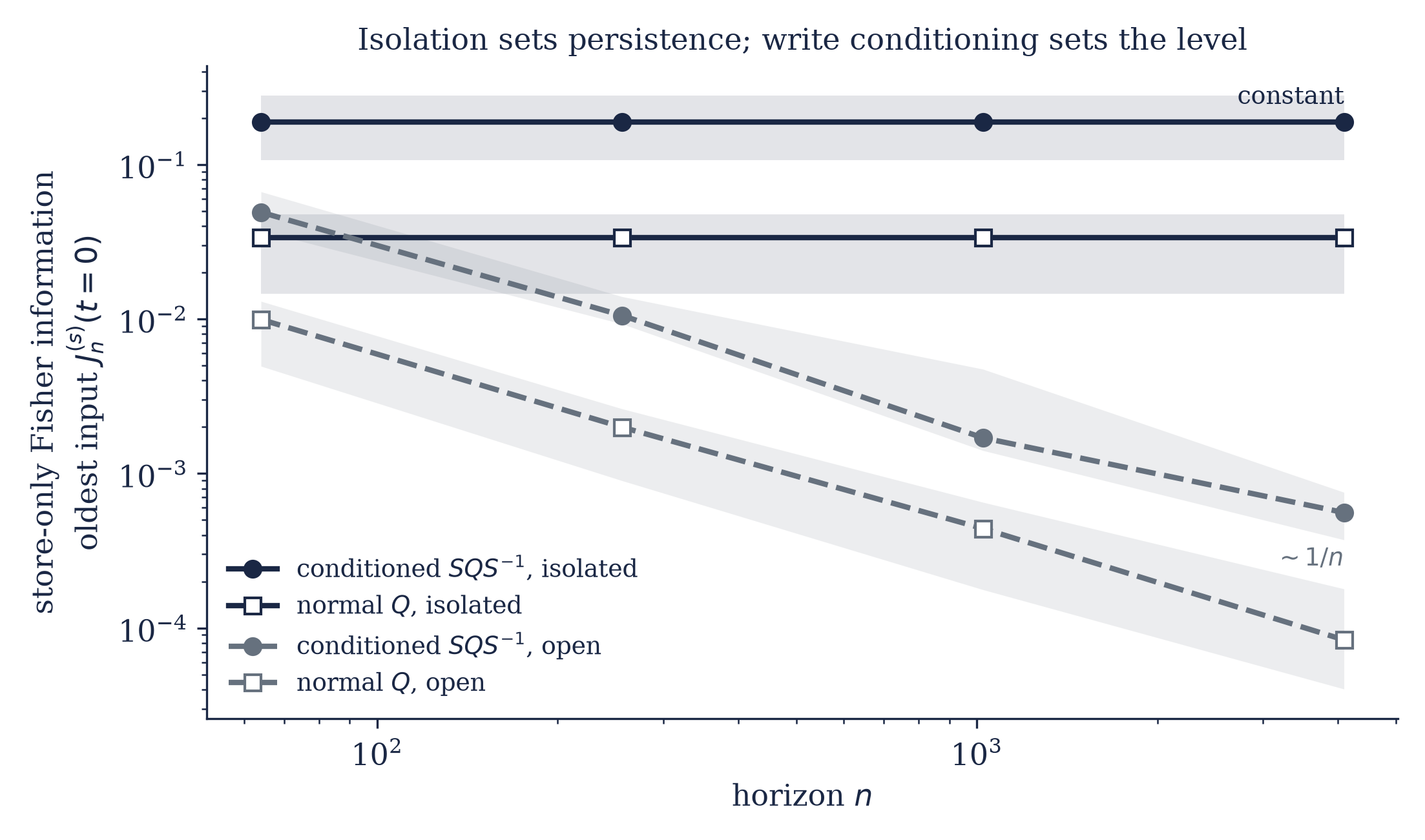}

\emph{Figure 4. Closing the coupling makes the oldest-input store
information constant for both writers. Leaving it open produces strong
decay. Conditioning changes the directional level, not the qualitative
horizon dependence.}

\subsubsection{5.3 Store-optimized direction
selection}\label{store-optimized-direction-selection}

For each of the same eight conditioned writers, we computed both \[
v_{\mathrm{write}}=\operatorname{eigmax}(M_{2048})
\] and \[
v_{\mathrm{store}}=\operatorname{eigmax}(M_{\mathrm{store}}).
\] The store trace was \(16\) to numerical precision in every draw. The
ratio \[
\frac{v_{\mathrm{store}}^\top M_{\mathrm{store}}v_{\mathrm{store}}}
{v_{\mathrm{write}}^\top M_{\mathrm{store}}v_{\mathrm{write}}}
\] had median \(1.224\) and range \(1.112\)--\(1.582\). The
corresponding oldest-input information ratio had median \(1.257\) and
range \(0.922\)--\(1.762\): maximizing the total caused one individual
loss on the oldest input. Every selected store direction still had
write-block allocation above \(1.05\) (median \(3.70\), range
\(3.11\)--\(4.29\)) and nonzero transfer of the designated oldest input.

End-to-end selection therefore improves the store-total objective by
construction, but it does not dominate every lag-specific objective. The
objective, together with any side constraints it needs, has to be fixed
before the direction is selected.

Stroud et al.~optimize working-memory loading for later retention
{[}37{]}, and Zylberberg et al.~show that most of their optimal upstream
noise covariances depend on downstream weights and noise {[}38{]}. These
are precedents for choosing an upstream configuration using its
downstream objective.

\subsection{6. Post-write retention and decoded
readout}\label{post-write-retention-and-decoded-readout}

\subsubsection{6.1 Exact post-write
invariance}\label{exact-post-write-invariance}

Let \(P_0\) contain closure-time sensitivities of stored writes and let
\(\Sigma_0\succ0\) be the closure covariance. If, after closure, no
further coupling or disturbance enters the store and the known hold is
invertible, \[
P_h=A_hP_0,\qquad \Sigma_h=A_h\Sigma_0A_h^\top,
\] then the full stored Fisher matrix is invariant: \[
\boxed{
P_h^\top\Sigma_h^{-1}P_h=P_0^\top\Sigma_0^{-1}P_0.
}
\] This holds for orthogonal, expanding and contracting invertible
holds. Contraction after closure is a change of coordinates in exact
covariance-aware arithmetic; contraction during writing changes what is
written.

\subsubsection{6.2 Approximate-isolation
guarantee}\label{approximate-isolation-guarantee}

Suppose the sensitivity remains \(p_h=A_hp_0\), while an additive
post-write disturbance contributes covariance \(R_h\) satisfying \[
0\preceq R_h\preceq \alpha_h A_h\Sigma_0A_h^\top.
\] Then \[
\Sigma_h=A_h\Sigma_0A_h^\top+R_h
\preceq
(1+\alpha_h)A_h\Sigma_0A_h^\top,
\] and inverse order gives \[
\boxed{
\frac{J_0}{1+\alpha_h}\le J_h\le J_0.
}
\] Consequently, a retained fraction \(\beta_{\mathrm{ret}}\) is
guaranteed whenever \[
\alpha_h\le\beta_{\mathrm{ret}}^{-1}-1.
\]

Two covariance choices attain the scalar lower bound. Proportional
contamination \(R_h=\alpha_h A_h\Sigma_0A_h^\top\) is multiplicative
covariance inflation, as used in ensemble filtering {[}39{]}.
Signal-aligned contamination \(R_h=\epsilon A_hp_0p_0^\top A_h^\top\),
with \(\epsilon\ge0\), attains it at \(\alpha_h=\epsilon J_0\), giving
\(J_h=J_0/(1+\epsilon J_0)\), the information-limiting-correlation
formula of Moreno-Bote et al.~{[}40, Eq. 5{]}. Appendix A.9 gives the
Sherman--Morrison derivation. The same bound applies to
nuisance-projected information. For nuisance sensitivity \(B_0\) and
target \(p_0\) define
\(J^{\text{eff}}_0=\min_b\,(p_0-B_0b)^\top\Sigma_0^{-1}(p_0-B_0b)\).
When target and nuisance sensitivities are transported by the same
invertible hold and the covariance inequality above holds, the
quadratic-form inequalities apply to every \(b\); minimizing over \(b\)
preserves the ordering, so
\(J^{\text{eff}}_h\ge J^{\text{eff}}_0/(1+\alpha_h)\).

The additive covariance form is an assumption: it requires no
cross-covariance with the closure state and no new parameter-dependent
mean. The Fisher interpretation assumes the Gaussian model; under a
covariance-only non-Gaussian specification the same algebra supports a
generalized-least-squares variance interpretation rather than a Fisher
one.

We tested the bound on eight stores at \(\alpha\in\{0.05,0.25,1\}\),
using both aligned disturbances \(R=\alpha\Sigma_0\), which attain
equality, and random positive-semidefinite disturbances bounded by
\(\alpha\Sigma_0\). All 48 checks passed; the minimum numerical margin
was \(-4.4\times10^{-16}\).

With independent white store noise of constant nonzero covariance
\(D\succeq0\) at each step and an orthogonal hold,
\(\|R_h\|_2\le h\|D\|_2\), so one may take
\(\alpha_h=h\|D\|_2/\lambda_{\min}(\Sigma_0)\). The resulting guaranteed
floor falls as \(1/h\); it is a lower bound, not an asserted loss rate.
Noise-driven diffusion of continuous-attractor memories is an analogous
accumulation mechanism {[}41{]}, under a different dynamical model.

Residual coupling is not covered by this inequality. It changes the
designated sensitivity, injects later inputs and write-path noise, and
can create cross-covariance with the closure state. Such a branch
requires full joint propagation and a recomputed decoder.

\subsubsection{6.3 Sampled generalized-least-squares
decoder}\label{sampled-generalized-least-squares-decoder}

For each draw we used \(v_{\mathrm{store}}\), stored one unknown scalar
input, generated 5,000 closure states, and applied the scalar GLS
decoder \[
a_0=\frac{\Sigma_0^{-1}p_0}{p_0^\top\Sigma_0^{-1}p_0},\qquad
\hat s=a_0^\top x_s.
\] Across the eight draws, the sampled error variance divided by the
theoretical value \(1/J_0\) had median \(1.017\) and range
\(0.969\)--\(1.032\); the sampled bias lay between \(-1.19\) and
\(1.04\) standard errors.

We then propagated the same realized states through \(U^H\) and
\(0.9^HU^H\) at \(H\in\{200,700\}\). All three readings reproduced the
closure estimate, with maximum absolute discrepancy
\(8.0\times10^{-15}\): (1) inverse-adjoint decoder propagation, (2)
state unwarping followed by the closure decoder, and (3) GLS
recomputation from propagated sensitivity and covariance.

This validates the scalar Gaussian readout; it does not establish joint
recovery of many unknown inputs or invariance of a fixed, quantized,
clipped or covariance-mismatched decoder.

\subsection{7. Task-trained realization and objective-specific
validation}\label{task-trained-realization-and-objective-specific-validation}

Sections 2 to 6 select the high-information write direction
analytically. We test whether behavioural training reaches the same
direction without access to the Fisher operator, and whether the
operator predicts the resulting behaviour. The released claim ledger
lists the source data for every number below.

\subsubsection{7.1 What was trained, what was held fixed, and how the
runs are
counted}\label{what-was-trained-what-was-held-fixed-and-how-the-runs-are-counted}

\textbf{Trainable:} an unconstrained vector \(u\in\mathbb R^{N}\), used
as the unit write direction \(v=u/\lVert u\rVert\), and a linear
logistic readout \((w,b)\).

\textbf{Held fixed:} the writer \(W\), the coupling \(K\), the store map
\(U\), the coupling schedule, the write interval and the query horizons.
No recurrent weight and no memory controller was learned; the
architecture is the one specified in §5.1 throughout.

Kang, Shirasaka and Suzuki {[}21{]} optimized input masks using the
leading eigenvector of a Fisher memory matrix and evaluated them on
memory tasks, for a contracting delay loop with an infinite-history
noise covariance. ROME {[}42{]} also derives task-weighted input
directions under a Frobenius-norm power constraint on the whitened
encoder and relates prediction-loss training of the encoder, with an
optimal linear readout, to that objective. Its task weights already
allow the designated delay to change. Here the comparison uses an exact
finite-window operator for a separate downstream store reached through
time-varying coupling, alongside the trace budget and post-closure
retention analysis. The training studies measure whether behavioural
optimization approaches that operator's task-specific optimum under a
fixed budget.

The optimizer receives the sensitivity map, the noise factor, the input
amplitude and the class labels. It never receives a Fisher matrix, an
eigenvector, a covariance inverse or an oracle target. An ablation
confirms this: a second trainer driven by the literal recurrence, which
never constructs the covariance, reaches the same direction.

Table 7.1 gives the full configuration, taken from the run records and
the calibration code rather than restated from the protocol.

\textbf{Table 7.1. Executed configuration.}

{\def\LTcaptype{none} 
\begin{longtable}[]{@{}
  >{\raggedright\arraybackslash}p{(\linewidth - 2\tabcolsep) * \real{0.5000}}
  >{\raggedright\arraybackslash}p{(\linewidth - 2\tabcolsep) * \real{0.5000}}@{}}
\toprule\noalign{}
\begin{minipage}[b]{\linewidth}\raggedright
field
\end{minipage} & \begin{minipage}[b]{\linewidth}\raggedright
value
\end{minipage} \\
\midrule\noalign{}
\endhead
\bottomrule\noalign{}
\endlastfoot
writer dimension \(N\), store dimension \(d\) & 32, 16 \\
write interval \(T_w\); designated input times \(t_0\) & 24; \(0\) and
\(12\) \\
training horizon \(H_{\text{train}}\); query horizons & 128;
\(64,256,1024,4096\) \\
input amplitude \(a\); class priors & \(1.5\); equal \\
initial state & \(x_w(0)=x_s(0)=0\) \\
noise schedule & writer innovations \(z_t\sim\mathcal N(0,I_N)\) enter
the write block at every step. Under \textbf{isolation} the coupling
closes at \(t=T_w=24\): only innovations inside the write interval reach
the store, and the endpoint is carried to the query horizon by
\(A=U^{H-T_w}\) with no further noise. Under \textbf{open coupling} the
same recursion runs to the query horizon, so write-path innovations keep
entering the store for all \(H\) steps. Disturbance added directly to
the store covariance occurs only in the declared approximate-isolation
conditions \\
optimizer & Adam, lr \(3\times10^{-3}\), weight decay \(0\), global
grad-norm clip \(1.0\) \\
batch size; steps; checkpoint & 512; 12,000; final \\
initialization & \(u\sim\mathcal N(0,I)\) then normalized; \(w=0\),
\(b=0\) \\
training sample & fresh draws each step from the endpoint law; no fixed
training set \\
calibration sample & 50,000 episodes, dedicated seed stream, disjoint
from test \\
test sample & 50,000 episodes per condition, dedicated seed stream \\
recalibrated readout & pooled-within-class discriminant fit from
calibration data; no analytic \(\Sigma\) enters it \\
fixed readout & the trained \((w,b)\) retained while only the direction
or horizon changes; analytic in neither case \\
decoder transport check & a decoder fitted at \(H_{\text{train}}\) in
its own right, by linear discriminant analysis on 50,000 sampled
endpoints, then transported by \(A^{-\top}\); not the trained \((w,b)\)
of the fixed-readout rows \\
direction controls & \(v_{\text{old}}\), \(v_{\text{store}}\),
\(v_{\text{write}}\), \(v_{\text{bottom}}\), \(v_{\text{worst}^+}\), 128
isotropic random, and the learned direction per seed \\
paired noise & within one (draw, writer type, horizon) every direction
is scored on the same standard normals and labels; the isolated and open
arms share them \\
aggregation order & median over the 5 seeds within a carrier first, then
the statistic across the 16 carriers \\
intervals & 10,000-resample paired bootstrap over carriers; one-sided
bounds are the 5th percentile of the median \\
secondary test & paired sign-flip permutation on the mean of
carrier-level differences, 10,000 resamples \\
\end{longtable}
}

\textbf{Counting.} The first study contains 160 optimization runs, the
second 320, and the replication in §7.4.1 a further 320 on a different
carrier block. These 800 runs are not 800 independent samples: the
independent unit is the carrier draw. Optimizer seeds are nested repeats
within a draw, the two objectives are paired within a draw, and the
normal and non-normal writers are paired constructions sharing \(Q\),
\(K\) and \(U\). Every interval reported here is a carrier-level paired
interval.

\(R_J\) is a normalized Fisher/Rayleigh efficiency in \([0,1]\), not a
classification accuracy: a value near \(0.999\) means the learned
direction captures that fraction of the leading eigenvalue of the task
operator.

\textbf{Independent implementations.} The primary path is float64 NumPy
on CPU. A second, independently written PyTorch implementation
reproduces the primary endpoint on four pilot cells to
\(7.0\times10^{-10}\) in \(R_J\). The literal sequential unroll and a
work-efficient parallel scan of the same recurrence are parity controls,
agreeing with the original sequential implementation to
\(4.6\times10^{-13}\) relative over 32 cells. These checks concern the
implementation, not the science; the accompanying throughput benchmarks
are in the supplement.

\subsubsection{7.2 Learned directional
efficiency}\label{learned-directional-efficiency}

With \(M_{\text{task}}\) the end-to-end operator for the designated
input, write

\[
R_J=\frac{v^\top M_{\text{task}}v}{\lambda_{\max}(M_{\text{task}})},
\qquad
A_{\text{sub}}=\bigl\lVert P_{\mathcal E_{0.99}}v\bigr\rVert_2^2 .
\]

{\def\LTcaptype{none} 
\begin{longtable}[]{@{}lrr@{}}
\toprule\noalign{}
& non-normal & normal \\
\midrule\noalign{}
\endhead
\bottomrule\noalign{}
\endlastfoot
median \(R_J\) & 0.99936 & 0.99891 \\
median \(A_{\text{sub}}\) & 0.99861 & 0.99614 \\
median \(R_J\) of an isotropic random direction & 0.1444 & 0.2621 \\
\end{longtable}
}

Medians are of per-carrier medians over 16 carriers, seeds nested.

The trained loss is a behavioural loss on the same designated input
whose information \(M_{\text{task}}\) measures, so the optimum of the
population loss over the write direction is the leading eigenvector of
\(M_{\text{task}}\). The finding is that a finite-budget optimizer that
is not told this optimum reaches it, not that a new optimum was found.

At a fixed write mask the logistic loss is convex in \((w,b)\), but the
joint problem over a unit-norm write mask and a readout is not: the
constraint set \(\{\lVert v\rVert=1\}\) is not convex, and the objective
is invariant under \((v,w,b)\mapsto(-v,-w,-b)\), so minimisers come in
pairs whose midpoint is infeasible and has strictly higher loss.

\subsubsection{7.3 Fisher-to-behaviour
calibration}\label{fisher-to-behaviour-calibration}

For an endpoint \(x\mid y\sim\mathcal N(y\,a\,p,\Sigma)\) with equal
priors, the Bayes-optimal linear rule attains
\(\operatorname{Acc}^\star=\Phi\!\left(a\sqrt J\right)\) with
\(J=p^\top\Sigma^{-1}p\); see Proposition A.11. This is an analytic
result about the model, and the experiment tests whether the
implementation realises it.

Over all evaluated directions, storage conditions and horizons, the mean
absolute difference between \(\Phi(a\sqrt J)\) and the empirical
accuracy of a covariance-aware readout refit on an independent
calibration sample is 0.00196 (non-normal) and 0.00189 (normal), pooled
over 22,080 rows each, with least-squares slopes 1.0068 and 1.0092
(Figure 5).

The agreement spans 4.23 orders of magnitude in \(J\), from
\(1.881\times10^{-5}\) to \(0.3171\), over every direction except the
null direction, whose rows are reported separately below.

\includegraphics[width=0.67\linewidth,height=\textheight,keepaspectratio]{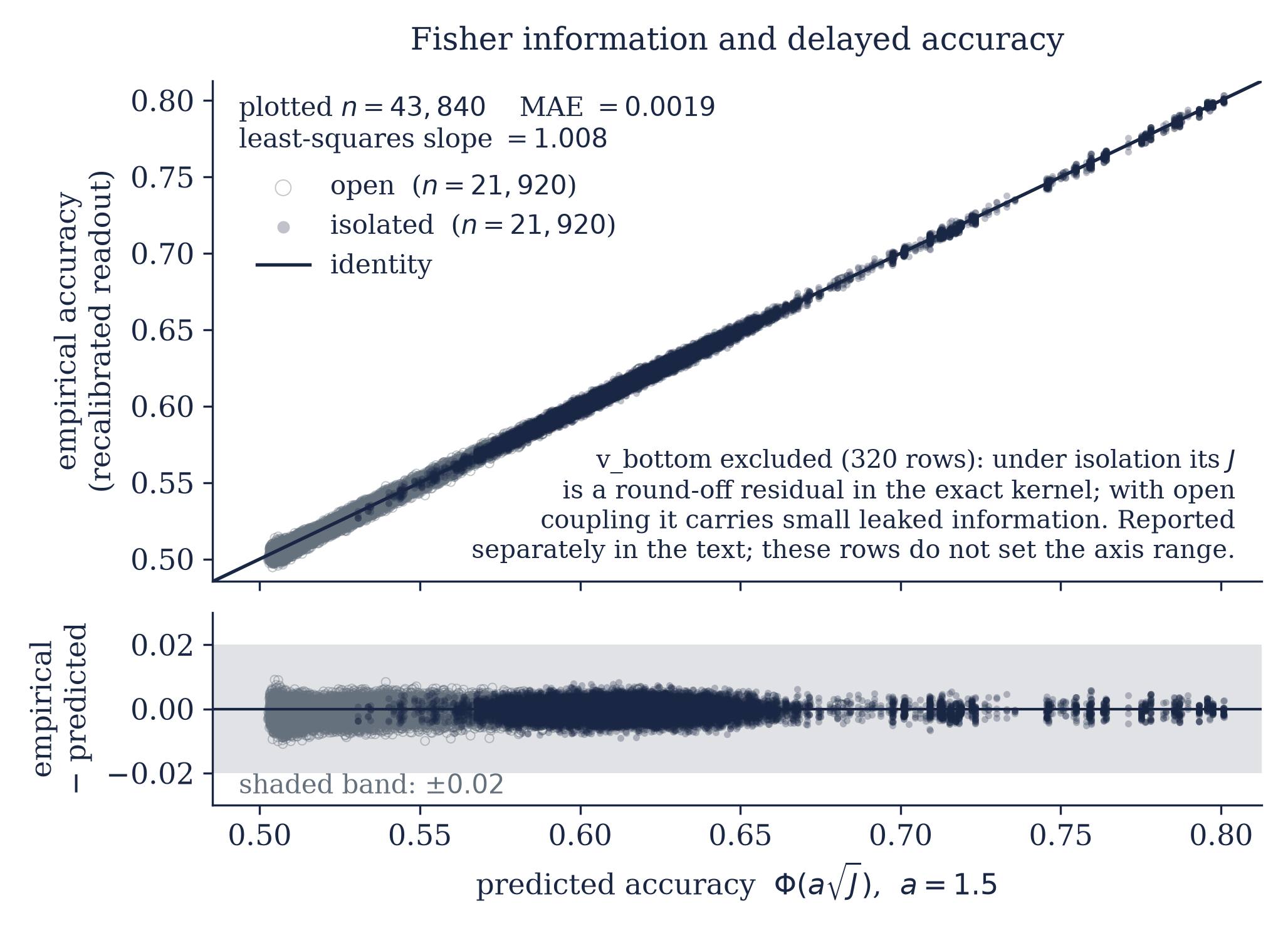}

\emph{Figure 5. Fisher-to-behaviour calibration. Predicted accuracy
\(\Phi(a\sqrt J)\) at \(a=1.5\) against the empirical accuracy of a
covariance-aware readout refit on an independent calibration sample, for
every evaluated direction, storage condition and horizon: 43,840 rows,
open and isolated drawn separately. The straight line is the identity,
not a fit. The lower panel is the residual, with a \(\pm 0.02\) band
drawn for scale only. The 320 \(v_{\text{bottom}}\) rows are excluded
from the plotted population and from the axis range, as explained in the
null-direction control.}

\emph{Null-direction control.} Because \(M_{\text{task}}\) has rank
\(d=16\) in an \(N=32\)-dimensional write space, its bottom eigenvector
lies in an exact 16-dimensional kernel of the isolated end-to-end
operator. Under isolation its information is \(J=0\) and its predicted
accuracy exactly \(1/2\); the computed values are round-off residuals,
with median \(1.86\times10^{-32}\) and maximum \(1.53\times10^{-30}\)
over the 160 isolated null-direction rows. With the coupling left open
the same direction receives leaked information, from
\(8.3\times10^{-6}\) to \(7.1\times10^{-3}\) over the corresponding 160
rows; that is not information in the kernel of the isolated operator. We
report the two populations separately, since a median taken across both
would fall between them and describe neither. Null-direction rows are
excluded from the reported range and from Figure 5 but remain in the
released data.

\subsubsection{7.4 Objective-specific geometry: a pre-specified
confirmation}\label{objective-specific-geometry-a-pre-specified-confirmation}

Changing which input the task designates changes \(M_{\text{task}}\). A
second, pre-specified study trained directions independently for two
designated input times, \(t_0=0\) and \(t_0=T_w/2=12\), on 16 carrier
draws from a seed range no earlier study had used:
\(16\times2\times5\times2=\mathbf{320}\) runs.

Write \(r_j(v)=v^\top M_j v/\lambda_{\max}(M_j)\). With \(v_j^\star\)
the oracle of objective \(j\),

\[
G^\star_j=1-r_j(v^\star_{\bar j}),\qquad
G^{\text{learn}}_j=r_j(v^{\text{learn}}_j)-r_j(v^{\text{learn}}_{\bar j}),\qquad
C_j=\frac{G^{\text{learn}}_j}{G^\star_j}.
\]

These carriers had already been run at reduced budget during
development, and those results had been seen before the confirmatory run
(§9.3). The study is therefore a pre-specified validation, not a blind
holdout confirmation.

{\def\LTcaptype{none} 
\begin{longtable}[]{@{}
  >{\raggedright\arraybackslash}p{(\linewidth - 8\tabcolsep) * \real{0.1579}}
  >{\raggedleft\arraybackslash}p{(\linewidth - 8\tabcolsep) * \real{0.2105}}
  >{\raggedleft\arraybackslash}p{(\linewidth - 8\tabcolsep) * \real{0.2105}}
  >{\raggedleft\arraybackslash}p{(\linewidth - 8\tabcolsep) * \real{0.2105}}
  >{\raggedleft\arraybackslash}p{(\linewidth - 8\tabcolsep) * \real{0.2105}}@{}}
\toprule\noalign{}
\begin{minipage}[b]{\linewidth}\raggedright
\end{minipage} & \begin{minipage}[b]{\linewidth}\raggedleft
\(t_0=0\) NN
\end{minipage} & \begin{minipage}[b]{\linewidth}\raggedleft
\(t_0=12\) NN
\end{minipage} & \begin{minipage}[b]{\linewidth}\raggedleft
\(t_0=0\) N
\end{minipage} & \begin{minipage}[b]{\linewidth}\raggedleft
\(t_0=12\) N
\end{minipage} \\
\midrule\noalign{}
\endhead
\bottomrule\noalign{}
\endlastfoot
median own-objective \(r_j(v^{\text{learn}}_j)\) & 0.99945 & 0.99945 &
0.99885 & 0.99874 \\
median available separation \(G^\star_j\) & 0.2393 & 0.2090 & 0.5095 &
0.4455 \\
median achieved separation \(G^{\text{learn}}_j\) & 0.2353 & 0.2126 &
0.5114 & 0.4412 \\
median contrast recovery \(C_j\) & 0.9990 & 1.0010 & 0.9997 & 0.9987 \\
\end{longtable}
}

\(G^{\text{learn}}_j>0\) in 16 of 16 carrier draws for both objectives
and both writer types, with every paired 95\% carrier-level bootstrap
interval excluding zero (Figure 6). The smallest available separation in
any cell was \(0.0603\), so no carrier was excluded as geometrically
indistinguishable.

The four contrasts are consequences of one successful optimization
rather than independent hypotheses, although none of them holds by
construction. The exact accounting is

\[
G^{\text{learn}}_j
= G^\star_j
-\bigl[1-r_j(v^{\text{learn}}_j)\bigr]
-\bigl[r_j(v^{\text{learn}}_{\bar j})-r_j(v^\star_{\bar j})\bigr],
\]

verified to a residual of \(0\) on all 64 cell-objective pairs. A
positive \(G^\star_j\) certifies that the two objectives \emph{permit}
separation; it does not certify what a finite optimizer will return. An
optimizer returning the same direction for both objectives would give
\(G^{\text{learn}}_j=0\) at any \(G^\star_j\). The second bracket is a
signed discrepancy, not a non-negative loss: it is negative wherever the
other objective's learned direction scores below that objective's own
oracle on \(M_j\). What the measured medians add is that both bracketed
terms are at or below \(1.3\times10^{-3}\) in magnitude, so the
optimizer converted essentially all of the available separation. We do
not derive a bound on either term by combining separately reported
medians.

\includegraphics[width=0.88\linewidth,height=\textheight,keepaspectratio]{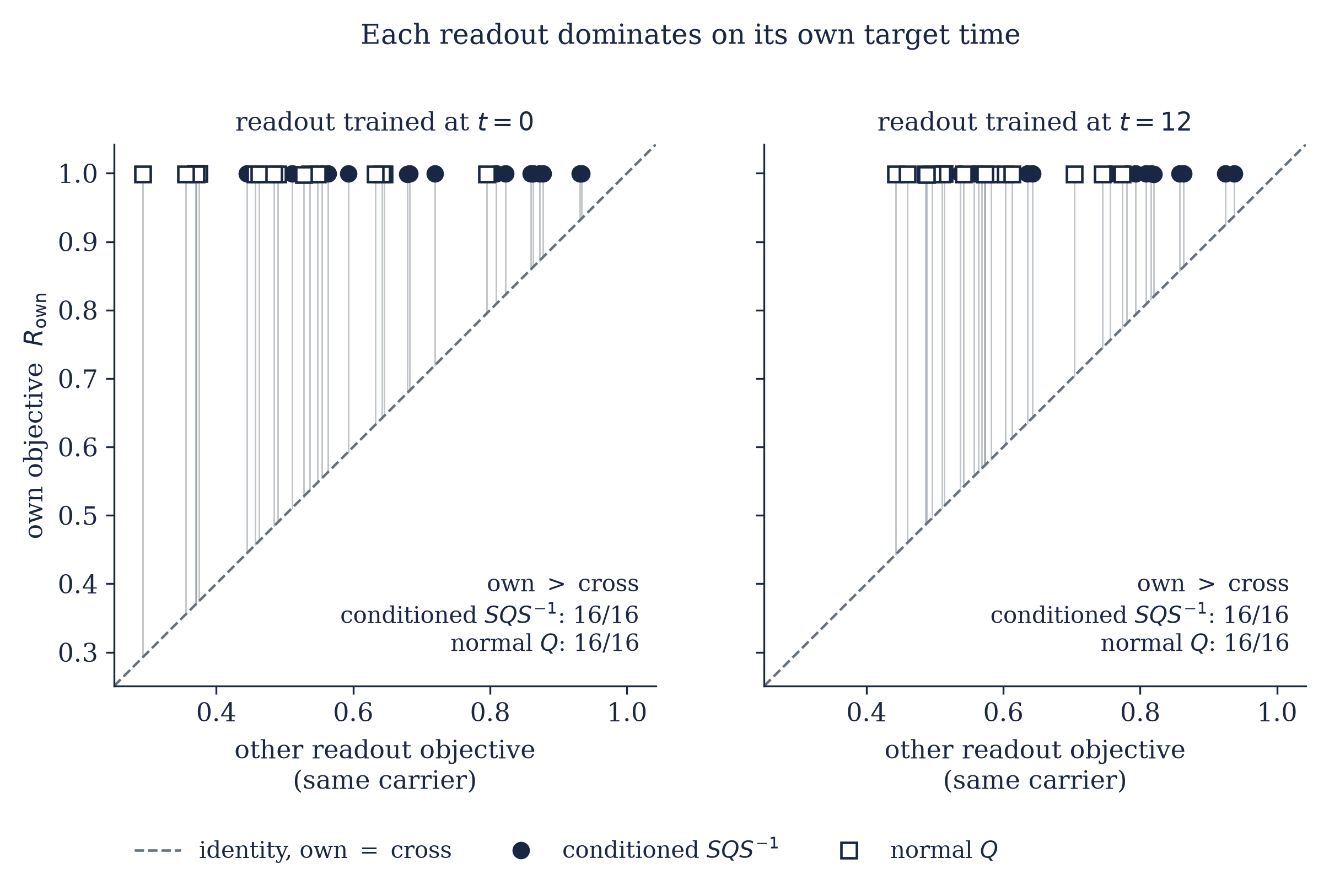}

\emph{Figure 6. Objective-specific geometry, one marker per carrier draw
and writer type. The vertical axis is \(r_j\) at the direction trained
for that panel's designated input time; the horizontal axis is the same
\(r_j\) evaluated at the direction trained for the other time, on the
same carrier. A marker above the dashed identity line is a carrier on
which the two objectives separated. The thin segment joins the two
values of each pair. The separation holds in 16 of 16 draws for both
writer types in both panels.}

\subsubsection{7.4.1 A second carrier block, unused before the
run}\label{a-second-carrier-block-unused-before-the-run}

Because the carriers above had been used during development, we ran the
identical design once more on a different block of draws, 16 to 31, that
no earlier study, development run, pilot, smoke test or benchmark had
instantiated.

We established that the block was unused before any training, from its
use history rather than from performance. A preflight compared the
block's generator identities and all 560 derived seed streams against
the 910 seeds reserved by the first study and against the 595 seeds of
every earlier draw of the second study, and compared carrier digests
against both populations. All four collision sets were empty, and no
requested draw had been used before. No reduced-budget training,
endpoint classification or oracle screening was run on the block, so no
carrier was kept or replaced because of how it performed. The block, its
seed streams, the code hashes and the execution environment were
recorded before the first model was trained.

The first execution on this block failed for a software reason and
returned 32 of 32 cells invalid. No validity check failed; one check
\emph{could not} fail. The negative branch of the seed-freshness check
had been written to trigger only when the audited draw index was
reserved by the first study, which holds for draws 0 to 15 but not for
16 to 31, so on this block it could not show the failure it is meant to
detect, and the preregistered rule then invalidated every cell. Because
invalid cells are discarded before any result is written, that execution
left no data. We corrected and reviewed the check, tested it only on
previously used draws, and ran the same design once more, having fixed
in advance that a second invalid execution would end the attempt without
a result.

The rerun produced 32 of 32 valid cells (0 invalid), 320 optimization
runs and 33 validity checks per cell, and it met every preregistered
criterion. We report it on its own and do not pool it with the first
block:

{\def\LTcaptype{none} 
\begin{longtable}[]{@{}
  >{\raggedright\arraybackslash}p{(\linewidth - 8\tabcolsep) * \real{0.1579}}
  >{\raggedleft\arraybackslash}p{(\linewidth - 8\tabcolsep) * \real{0.2105}}
  >{\raggedleft\arraybackslash}p{(\linewidth - 8\tabcolsep) * \real{0.2105}}
  >{\raggedleft\arraybackslash}p{(\linewidth - 8\tabcolsep) * \real{0.2105}}
  >{\raggedleft\arraybackslash}p{(\linewidth - 8\tabcolsep) * \real{0.2105}}@{}}
\toprule\noalign{}
\begin{minipage}[b]{\linewidth}\raggedright
\end{minipage} & \begin{minipage}[b]{\linewidth}\raggedleft
\(t_0=0\) NN
\end{minipage} & \begin{minipage}[b]{\linewidth}\raggedleft
\(t_0=12\) NN
\end{minipage} & \begin{minipage}[b]{\linewidth}\raggedleft
\(t_0=0\) N
\end{minipage} & \begin{minipage}[b]{\linewidth}\raggedleft
\(t_0=12\) N
\end{minipage} \\
\midrule\noalign{}
\endhead
\bottomrule\noalign{}
\endlastfoot
median own-objective \(r_j(v^{\text{learn}}_j)\) & 0.9994 & 0.9995 &
0.9989 & 0.9987 \\
median available separation \(G^\star_j\) & 0.2584 & 0.2922 & 0.4411 &
0.4581 \\
median achieved separation \(G^{\text{learn}}_j\) & 0.2597 & 0.2904 &
0.4509 & 0.4613 \\
median contrast recovery \(C_j\) & 1.0005 & 0.9984 & 0.9975 & 0.9969 \\
\end{longtable}
}

\(G^{\text{learn}}_j>0\) in 16 of 16 carrier draws for both objectives
and both writer types, with every paired 95\% carrier-level bootstrap
interval excluding zero. The smallest available separation in any cell
was \(0.0803\) and no cell was flagged negligible, so all 16 carriers
were usable in each condition.

This replication shows the same finite-budget optimization result on
carriers whose outcomes were unknown when the run was committed, which
the first block could not show. It does not widen the scope of the
paper: the systems, the training setup, the objectives and the
limitations of §9 are unchanged, and the first block is still reported
as a pre-specified validation.

\subsubsection{7.5 Allocation and retention under storage
intervention}\label{allocation-and-retention-under-storage-intervention}

All results in this subsection use the same learned direction, with no
retraining between conditions.

Under exact isolation with an invertible hold the stored Fisher
information is invariant in the horizon, and the measurement reproduces
this: the pooled median \(J\) is 0.2553 (non-normal) and 0.1406
(normal), identical to every reported digit across \(H=64\) to \(4096\)
(Figure 7).

Two different ratios describe continued coupling. Each is formed per
carrier from that carrier's seed medians, so the seed median is taken
before the ratio, and the tabulated value is the median of those
per-carrier ratios for each writer type, never pooled across writer
types:

{\def\LTcaptype{none} 
\begin{longtable}[]{@{}
  >{\raggedright\arraybackslash}p{(\linewidth - 4\tabcolsep) * \real{0.2727}}
  >{\raggedleft\arraybackslash}p{(\linewidth - 4\tabcolsep) * \real{0.3636}}
  >{\raggedleft\arraybackslash}p{(\linewidth - 4\tabcolsep) * \real{0.3636}}@{}}
\toprule\noalign{}
\begin{minipage}[b]{\linewidth}\raggedright
comparison
\end{minipage} & \begin{minipage}[b]{\linewidth}\raggedleft
non-normal
\end{minipage} & \begin{minipage}[b]{\linewidth}\raggedleft
normal
\end{minipage} \\
\midrule\noalign{}
\endhead
\bottomrule\noalign{}
\endlastfoot
open store, \(J(H{=}64)/J(H{=}4096)\), decline along the horizon & 118.3
& 119.3 \\
\(J(\text{isolated})/J(\text{open})\), both at \(H=4096\), isolated
versus open at a fixed horizon & 520.4 & 486.9 \\
\end{longtable}
}

Ratios of pooled medians are a different estimand (\(130.5\) and
\(133.6\) along the horizon; \(580.7\) and \(520.4\) for the
fixed-horizon contrast) and are given in the supplement with their
interquartile ranges. In particular, the 580-fold figure is a contrast
at a fixed horizon, not a decline from \(H=64\) to \(H=4096\).

Approximate isolation with an additive disturbance bounded by \(\alpha\)
times the closure covariance satisfied its guaranteed floor
\(J_h\ge J_0/(1+\alpha)\) in all 4,800 checked rows, with the aligned
construction attaining the bound exactly, as the theorem requires.

Invertibility is a sufficient condition for preservation of the complete
Fisher matrix for arbitrary stored sensitivities under the stated hold.
It is not necessary for a designated target: for
\(x\mid s\sim\mathcal N\bigl((s,0)^\top,I_2\bigr)\) the scalar
information is one, and the non-invertible projection \(x\mapsto x_1\)
retains it exactly, discarding only a parameter-independent coordinate.
More generally a non-invertible map may preserve a sufficient statistic
for the designated targets. The non-invertible control tested here, a
rank-8 projection of the 16-dimensional store, did discard informative
components and reduced the measured information in every cell (median
retained fraction \(0.47\)). That is a property of the tested
projection, not of non-invertibility as such.

\includegraphics[width=1\linewidth,height=\textheight,keepaspectratio]{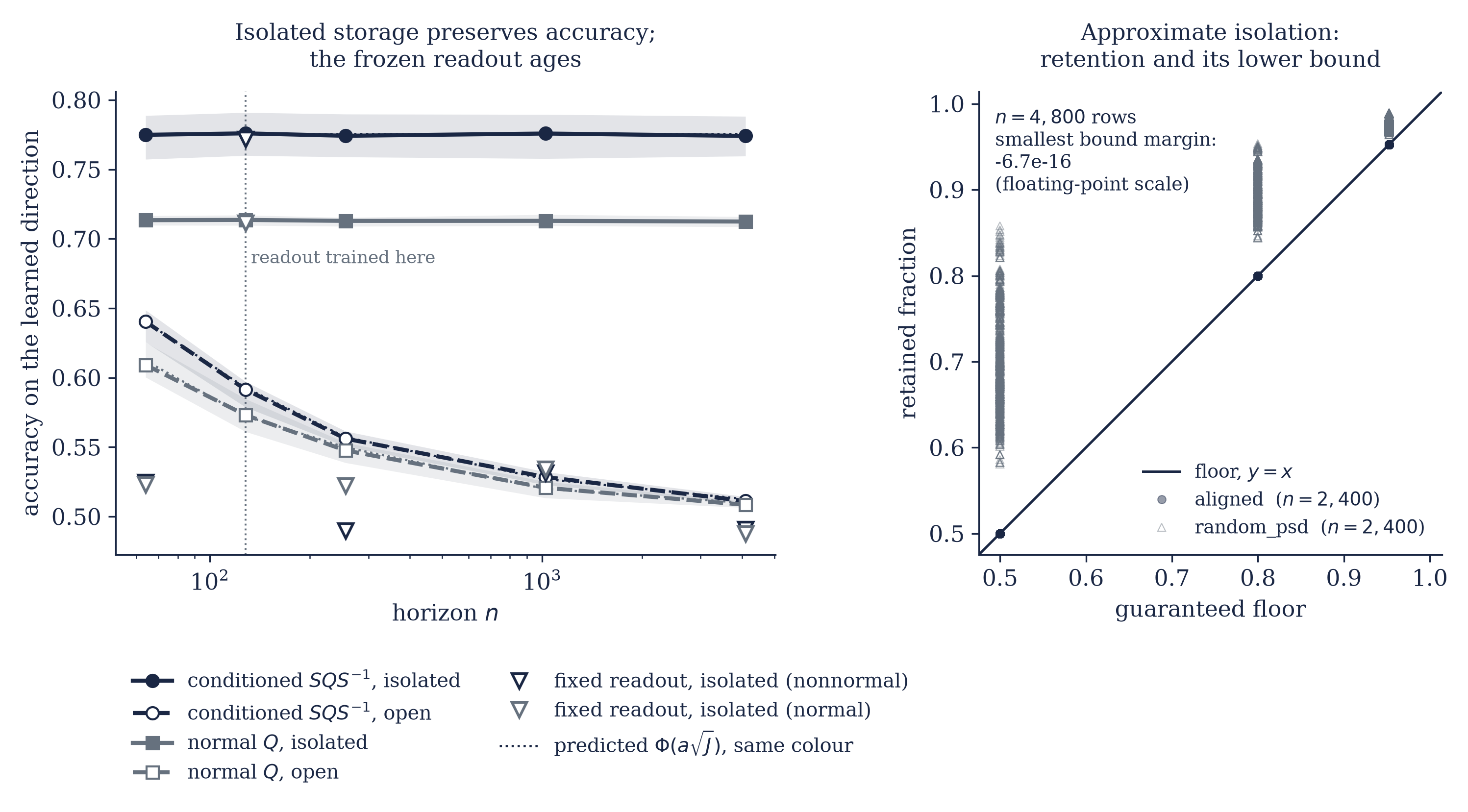}

\emph{Figure 7. Storage intervention and decoder transport. Left:
accuracy on the learned direction against the horizon, medians over 16
carrier draws and 5 seeds, with interquartile bands over those 80 rows
per cell. Closing the coupling holds the accuracy flat for both writer
types; leaving it open decays toward chance. Dotted overlays are the
predicted \(\Phi(a\sqrt J)\) in the same colour. The triangles are the
single readout \((w,b)\) that behavioural training produced at
\(H_{\text{train}}=128\), reused at every query horizon under isolation
with no retraining and no transport: it meets the isolated curve where
it was fitted and sits near chance at every other horizon, while the
stored information is unchanged (Section 7.6). They are left unjoined so
that the near-vertical excursion at \(H_{\text{train}}\) is not read as
a trajectory. Right: approximate isolation against its guaranteed floor
\(J_h\ge J_0/(1+\alpha)\), 4,800 rows; the aligned construction lies on
the bound and the random positive-semidefinite disturbances lie above
it.}

\subsubsection{7.6 Decoder aging is coordinate drift, not information
loss}\label{decoder-aging-is-coordinate-drift-not-information-loss}

Between the training horizon and a later query horizon the store applies
a known invertible map \(A=U^{\,H-H_{\text{train}}}\). Exact Fisher
information is preserved. A decoder fitted at \(H_{\text{train}}\)
remains tied to its training coordinate frame.

{\def\LTcaptype{none} 
\begin{longtable}[]{@{}
  >{\raggedright\arraybackslash}p{(\linewidth - 8\tabcolsep) * \real{0.1579}}
  >{\raggedleft\arraybackslash}p{(\linewidth - 8\tabcolsep) * \real{0.2105}}
  >{\raggedleft\arraybackslash}p{(\linewidth - 8\tabcolsep) * \real{0.2105}}
  >{\raggedleft\arraybackslash}p{(\linewidth - 8\tabcolsep) * \real{0.2105}}
  >{\raggedleft\arraybackslash}p{(\linewidth - 8\tabcolsep) * \real{0.2105}}@{}}
\toprule\noalign{}
\begin{minipage}[b]{\linewidth}\raggedright
\end{minipage} & \begin{minipage}[b]{\linewidth}\raggedleft
fixed decoder at \(H_{\text{train}}\)
\end{minipage} & \begin{minipage}[b]{\linewidth}\raggedleft
covariance-aware at \(H_{\text{train}}\)
\end{minipage} & \begin{minipage}[b]{\linewidth}\raggedleft
fixed decoder away
\end{minipage} & \begin{minipage}[b]{\linewidth}\raggedleft
covariance-aware away
\end{minipage} \\
\midrule\noalign{}
\endhead
\bottomrule\noalign{}
\endlastfoot
non-normal & 0.7718 & 0.7760 & 0.5029 & 0.7746 \\
normal & 0.7114 & 0.7136 & 0.5136 & 0.7128 \\
\end{longtable}
}

Pooled medians over 16 carriers and 5 seeds, and over the four
non-training horizons for the ``away'' columns.

The fixed decoder works where it was fitted and falls to chance
elsewhere while the information is provably unchanged. The remedy is a
change of coordinates, and it applies to any decoder tied to the
training frame. The transport check uses a separate decoder, not the
trained \((w,b)\): one fitted at \(H_{\text{train}}\) by linear
discriminant analysis on 50,000 sampled endpoints. Because the identity
holds for any decoder, this is a second instance of the same statement
rather than a reuse of the trained coefficients. Transporting that
decoder by the inverse adjoint, \(w_H=A^{-\top}w\), restores it: on the
same sampled episodes carried through the known hold, the transported
decoder reproduces the training-horizon logit to a worst absolute
difference of \(7.99\times10^{-15}\) and the training-horizon accuracy
with restoration error \(0\), across 32 rows (Figure 7).

A decoder can become obsolete without any loss of stored information.
The pointwise transport check verifies this distinction in the specified
linear-Gaussian model; biological memory and forgetting in general
neural systems remain outside the tested scope.

Temporal generalization tests whether a decoder transfers across time
{[}43{]}; dynamic activity can also support a stable coding subspace
{[}44{]}, and representational drift can leave fixed linear classifiers
effective {[}45{]}. Degenhart et al.~estimate an alignment between
neural activity spaces to stabilize a brain-computer interface {[}46{]},
while Joshi et al.~use measured hardware responses and recalibrated
batch-normalization statistics to compensate conductance drift {[}47{]}.
Those compensation procedures require data to estimate the correction;
the invertible hold map used here is known.

\subsubsection{7.7 Evidence taxonomy}\label{evidence-taxonomy}

Validity-check histories and performance benchmarks are in the
supplement.

{\def\LTcaptype{none} 
\begin{longtable}[]{@{}
  >{\raggedright\arraybackslash}p{(\linewidth - 4\tabcolsep) * \real{0.3333}}
  >{\raggedright\arraybackslash}p{(\linewidth - 4\tabcolsep) * \real{0.3333}}
  >{\raggedright\arraybackslash}p{(\linewidth - 4\tabcolsep) * \real{0.3333}}@{}}
\toprule\noalign{}
\begin{minipage}[b]{\linewidth}\raggedright
category
\end{minipage} & \begin{minipage}[b]{\linewidth}\raggedright
items in this section
\end{minipage} & \begin{minipage}[b]{\linewidth}\raggedright
what a reader may conclude
\end{minipage} \\
\midrule\noalign{}
\endhead
\bottomrule\noalign{}
\endlastfoot
standard identities & \(\operatorname{tr}M_n=N\); \(M_n=I\) for a normal
carrier; \(\operatorname{Acc}^\star=\Phi(a\sqrt J)\); post-write Fisher
invariance under an invertible hold & properties of the model,
established analytically \\
derived theorem claims & store trace bound; approximate-isolation floor
\(J_0/(1+\alpha)\); the \(G^{\text{learn}}\) accounting identity &
proved in the appendices, verified numerically \\
numerical implementation tests & sequential-versus-block parity; scan
parity; CPU-versus-CUDA endpoint parity; empirical-versus-analytic
covariance and Fisher; decoder transport; calibration error and slope &
the implementation realises the analytic results to its stated numerical
and statistical accuracy \\
finite-budget optimization results & \(R_J\), \(A_{\text{sub}}\),
own-objective \(r_j\), \(G^{\text{learn}}\), \(C_j\) & behavioural
training reaches the analytic optimum under the stated budget \\
exploratory observations & the preliminary target-time probe;
preliminary oracle separations; runtime benchmarks &
hypothesis-generating only, not confirmatory \\
\end{longtable}
}

The validity-check counts of the two studies are not independent
scientific hypotheses. The Holm-adjusted \(p\)-values of \(0.0004\) and
\(0.0008\) sit at the resolution floor of 10,000 resamples and do not
represent eight separate discoveries.

\subsection{8. Implications: allocation, admission and
retention}\label{implications-allocation-admission-and-retention}

The results provide separate descriptions of information available in
the writer, information admitted to the store, and information
accessible at readout. Each stage has its own objective and channel
constraints; a gain at one stage must be checked at the next.

\begin{enumerate}
\def\labelenumi{\arabic{enumi}.}
\item
  \textbf{Allocation.} The write-block operator \(M_n\) or \(M_\infty\)
  describes how a fixed trace budget is distributed over write
  directions. Normal carriers are isotropic. Non-normal carriers can
  place more information in selected directions at the expense of
  others.
\item
  \textbf{Admission.} The end-to-end operator \(M_{\mathrm{store}}\)
  includes the coupling and closure covariance. A direction selected by
  the write-block oracle may be suboptimal at the store. Selection
  should use the required store objective while retaining a
  write-allocation and transfer constraint.
\item
  \textbf{Retention and readout.} Exact isolation preserves the complete
  Fisher matrix under an invertible hold. Bounded additive contamination
  gives a quantitative retained-information floor. A fixed decoder can
  nevertheless lose accuracy as coordinates change. Residual coupling,
  observation noise, non-invertible channels, quantization and finite
  precision require separate accounting.
\end{enumerate}

Write geometry changes the direction and level of admitted information.
The post-write channel determines whether that geometry is preserved,
contaminated or discarded. These roles are separable, but their
magnitudes are not independent: coupling and storage choices can also
change the absolute information level. The effects therefore do not
combine multiplicatively in general.

\subsection{9. Limitations and open
experiments}\label{limitations-and-open-experiments}

All systems are linear and synthetic. The directional census uses
\(N=32\), eight instances per configuration, one process-noise model and
a post-hoc oracle direction. The store experiments use one conditioning
level, one coupling family, a 16-dimensional store and one scalar
conditioned-input decoder. The log relationship between transient gain
and oracle capacity is exploratory and family-specific.

The finite-horizon certificate can be conservative; the direct \(M_n\)
cross-check remains necessary when the spectral-group gap is small. The
end-to-end store oracle optimizes a total and does not guarantee
improvement for every lag. The approximate-isolation theorem assumes an
additive covariance order and zero residual coupling. The sampled
decoder assumes a correctly specified covariance and one unknown scalar.

Sections 7.2 to 7.4 establish finite-budget optimization of input masks
and linear readouts on fixed linear-Gaussian recurrent carriers. They do
not establish that learned recurrent weights, a learned coupling or
memory controller, nonlinear models, or systems with an unknown noise
law acquire the same geometry. Downstream utility beyond the
binary-estimation task studied here remains untested, and a Fisher-level
advantage can coexist with poor decoded accuracy when the absolute
signal-to-noise ratio is too small.

\subsubsection{9.1 Scope of the training
studies}\label{scope-of-the-training-studies}

The trained systems are the same linear, Gaussian memories analysed in
sections 2 to 6. Only the input mask and a linear readout were trained;
the writer, the coupling, the store map and the coupling schedule were
held fixed throughout. Extension to nonlinear systems, language models,
continual learning and catastrophic forgetting remains untested. These
experiments do not establish a ranking against modern sequence
architectures.

Both writer types met every criterion of the confirmation, but their
own-objective efficiencies and task-specialization gaps are computed
against \emph{different} task operators, so they do not show that
non-normal writers outperform normal writers on any task. The normal
write-block identity \(M_n=I\) concerns the sum over lags within the
write block; it does not imply that a lag-specific end-to-end store
operator is isotropic, and the measured store geometry of a normal
carrier is not isotropic.

With one optimizer, one step count and one amplitude, behavioural
training approached the task-specific analytic optimum. Calibration and
transport measurements agree with the model's analytic predictions.
General laws of neural memory remain outside the scope of these
experiments.

Several preregistered decision-rule items in the first study are
algebraic identities of the model rather than falsifiable hypotheses,
and are reported as identity checks. The confirmation contributes one
confirmatory optimization result with related consequences, not four
independent discoveries.

\subsubsection{9.2 Sampling from the endpoint
distribution}\label{sampling-from-the-endpoint-distribution}

Under the model's assumptions the store state at a query horizon is
exactly Gaussian, so episodes were drawn from that endpoint distribution
rather than by unrolling the recurrence. This is equivalent in
distribution to running the linear recurrence, and it does not mean that
recurrent weights were trained. A literal sequential unroll is kept as a
parity check in the released code.

\subsubsection{9.3 Provenance of the confirmation's carrier
draws}\label{provenance-of-the-confirmations-carrier-draws}

The second study used carrier draws disjoint from the earlier studies,
checked over all 560 seeds the run draws. The same draws had, however,
been run at reduced budget during development, and their summary results
had been inspected. The original eleven numerical thresholds and the two
target objectives were not changed; two additional, documented rules
tightened eligibility but were not triggered in the final run. The
detailed outputs of one earlier development run were overwritten and are
unavailable. Different QR output bytes under different BLAS
implementations do not make those carrier draws independent.

We therefore report this study as a pre-specified validation on carriers
seen during development, not as a blind holdout confirmation, and the
later block does not change that. The replication in §7.4.1 is a
separate study, whose carriers were unused before the run and whose one
invalid execution, which left no data, is reported with it. The records
show which thresholds were fixed in advance; they cannot exclude every
outcome-informed choice made during implementation or in deciding when
to proceed. The full disclosure is included in the release.

\subsection{10. Related work}\label{related-work}

\textbf{Fisher memory of linear carriers.} Ganguli, Huh and Sompolinsky
{[}1{]} introduced the FMC with in-loop noise and proved, under
stability, that normal carriers have total capacity one and any carrier
at most \(N\), with extensive capacity reached only by strongly
non-normal feedforward constructions. Orhan and Pitkow {[}3{]} restate
the normal-matrix result and reach order-\(N\) capacity with decaying
non-normal constructions; Hennequin, Vogels and Gerstner {[}2{]} give
the variance-amplification analogue (normal \(\Rightarrow\) no transient
amplification). Asllani, Lambiotte and Carletti {[}4{]} relate
non-normal network structure to transient amplification in linearly
stable systems. Tiňo {[}12{]} proves, for symmetric contracting Wigner
carriers, that memory over \(k\ge1\) is maximized by writing along the
dominant eigenvector, the precedent for our write-direction question,
with a different mechanism (slowest normal mode versus redistribution of
a fixed trace). Kerg et al.~{[}5{]} parameterize a broad Schur class
with unit or near-unit eigenspectra and non-orthogonal eigenbases, with
orthogonal matrices as a subset. Their reported Fisher-memory
calculations (their Proposition 1 and Table 6, \(J_{tot}\) from 3.0 to
20.5) focus on strictly lower-triangular chains with optional diagonal,
i.e.~nilpotent or contracting examples. We did not locate in that paper
a Fisher-memory characterization that isolates the finite-dimensional
bi-power-bounded non-normal subclass studied here; we do not
characterize their full parameterization as non-bi-power-bounded. Kang,
Shirasaka and Suzuki {[}21{]} take the write-direction rule of {[}1{]},
the top eigenvector of the Fisher memory matrix, and use it to optimize
the input mask of a Mackey--Glass time-delay reservoir under white state
noise. Their carrier is a contracting delay loop. They note that the
mask changes the performance of a fixed capacity, not the capacity
itself, which is the trace budget of §3.1 seen from the reservoir side.

The noise model and information functional must be specified together.
Here isotropic in-loop noise follows the same propagator as the signal,
fixing the spatial Fisher trace. Baggio et al.~{[}32{]} instead maximize
Gaussian mutual information over input covariances at fixed power, with
additive receiver noise and interference from previous packets; their
directed-chain examples gain capacity through non-normal amplification.
That optimized log-determinant capacity can rise while the Fisher trace
of the separate in-loop model remains fixed: neither quantity is the
other model's capacity. Jaeger's state-noise experiments show sharply
reduced linear-reconstruction memory {[}15{]}, while Guan et
al.~{[}18{]} obtain a noise-spectrum dependence when signal and noise
share an input channel.

ROME {[}42{]} constructs a task-weighted operator from linear responses
and a reference fluctuation covariance, then optimizes input encoding
under a power constraint. It reports encoder-training agreement in a
linear reservoir and tests nonlinear echo-state, spin-wave and spiking
reservoirs; its fixed-metric approximation can deteriorate at stronger
input. Its delay-dependent objectives already cover task-specific
direction selection. Here we also study the finite-window
downstream-store operator, its trace constraint, and the post-write
retention channel.

\textbf{Reconstruction memory and its completeness identities.} Jaeger
{[}15, 17{]} defines the reconstruction memory capacity and proves
\(\mathrm{MC}\le N\), with equality iff the Krylov matrix of the write
vector has full rank. In noise experiments he attributes the collapse of
memory under state noise to iterates \(W^k\) collapsing onto a
low-dimensional subspace and forcing large output weights, which
near-unitary \(W\) avoids. This is non-normal transient geometry seen
from the reconstruction side, where it destroys capacity; the Fisher
functional sees it as concentration of a fixed trace. White, Lee and
Sompolinsky {[}13{]} use the same in-loop dynamics with a
signal-plus-noise covariance; their Eq. (4), Dambre et al.'s
completeness theorem {[}19, Thm. 7{]} and Guan et al.'s \(M_{sum}=N\)
{[}18, supp. Eq. 108{]} are the reconstruction-side cousins of §3.1's
trace budget. In White et al.~orthogonal carriers are extensive with an
optimum just below exact reversibility, and random Gaussian carriers are
not; the two functionals differ in where the covariance puts the signal,
not in the noise model. Hermans and Schrauwen {[}16{]} show that without
noise the reconstruction memory function depends only on the eigenvalues
of \(W\) and is similarity-invariant, the opposite of the in-loop-noise
Fisher setting, where similarity by \(S\) is exactly what moves capacity
between directions. Haruna and Nakajima {[}10{]} bound the
reconstruction memory function from below by a harmonic memory
\(h(k)=\|p_k\|^2/(p_k^\top Cp_k)\), a Cramér--Rao form, with equality
iff the eigenvalues of the state covariance are constant on the support
of the write direction's projection. They observe that for random
Gaussian (non-normal) reservoirs the inequality is strict because the
noise and signal covariances do not share eigenvectors. That is a
non-normality signature in the reconstruction functional and the nearest
published statement to §3.1's directional spreading. Guan et
al.~{[}18{]} show that memory lost to input-channel noise is fixed by
the noise power spectrum.

\textbf{Persistence, stability and marginal dynamics.} Toyoizumi and
Abbott {[}14{]} find that memory lifetime diverges at the edge of chaos
only without internal noise. The curse-of-memory results {[}8{]} and
reversible RNNs {[}7{]} state, for different reasons, that stable
approximation forces decay and that reversible networks cannot forget;
UnICORNN {[}9{]} is a working time-invertible Hamiltonian recurrent
network with no capacity statement. Goldman {[}6{]} obtains memory
without feedback from purely feedforward (nilpotent) structure, the
extensive-capacity construction in its cleanest form. In the sources
reviewed, we did not locate the combination of the bi-power-bounded
limit theorem, finite-horizon certification, end-to-end store-direction
selection, and post-write retention guarantees stated here. The
invariance theorem itself is a standard property of Fisher information
under invertible transformations; the contribution is its placement
inside an allocation--admission--retention design and certification
chain.

\textbf{Covariance-tracking memories and readouts.} The readout of
§§5--6 propagates the stored mean and covariance through the
post-closure map and reads the Fisher form; these are the standard
operations of Kalman covariance propagation, and the invariance
statement is itself standard: Fisher information is preserved under a
known, parameter-independent invertible transformation of the
observation. We also measure how a fixed implemented decoder responds to
the known post-write map. Becker et al.~{[}25{]} propagate a factorized
latent covariance inside a recurrent network for uncertainty-aware
fusion. Dowling, Jeon, Savin and Park {[}24{]} derive recurrent layers
from an explicit memory design model. Their linear-Gaussian Bayesian
Layer propagates mean and covariance, and uses the covariance to steer
writes toward uncertain directions and protect confident ones. It
recovers linear attention, gated linear attention (GLA) and Mamba-2 as
exact filters, and DeltaNet as a covariance-reset reduction. It is a
close architecture-level neighbor because it also makes covariance part
of the memory state. Its purpose is Bayesian filtering and
uncertainty-aware writing; our purpose is to characterize a
trace-constrained directional Fisher geometry, propagate it through a
specified coupling, and certify the post-write channel. Fast weights
{[}22{]} and xLSTM {[}23{]} expose a learning rate and a decay rate as
separate controls, which is the pair the §6.1--6.2 analysis addresses.

Titans {[}27{]} likewise separates the gradient-update scale from a
multiplicative memory-forgetting factor. Beuria and Shukla {[}26{]}
build a reservoir from exactly discretized damped rotations, so that
rotation and decay are separate design variables and the operator is
normal by construction. By the normal-isotropy theorem of §3.2 such a
reservoir allocates exactly one unit of Fisher information to every
write direction at every horizon. Its separation is spectral, not the
write-path/storage-path separation of this paper, and it cannot
concentrate.

\textbf{Black-box Fisher-information-rate estimation.} Shi and Rojas
{[}28{]} estimate the Fisher information rate of a process with memory
from simulator output by combining local KL-divergence curvature,
context-tree weighting and least-squares matrix recovery. Their
parameter-Fisher information rate is a different object from the
past-input Fisher memory studied here. The two approaches are
complementary. Their method estimates an information geometry from
simulator output; we derive and certify the geometry of a specified
linear recurrent memory and use it to choose write and store directions.

\textbf{Projection and memory kernels.} Wang, Benner and Heiland
{[}20{]} derive, for a partially observed linear time-invariant system
with \(P f(x_1,x_2)=f(x_1,0)\), the closed-form Mori--Zwanzig
decomposition into Markovian term \(A_{11}x_1\), noise term
\(A_{12}e^{tA_{22}}x_2(0)\) and memory kernel
\(K(s)=A_{12}e^{sA_{22}}A_{21}\), and note it coincides with the
variation-of-constants formula. Their closed kernel is closely related
mathematically to the read--transport--write maps used here. We use the
specified recurrent memory to study directional Fisher allocation,
store-operator selection and post-write certification; we do not
introduce a new elimination identity.

Bai, Li and Kou {[}48{]} obtain a projector-rank sum rule for spatial
Fisher-information retention in non-Hermitian feed-forward chains under
a fixed internal quadratic resource; directionality redistributes access
among ports. Wang and Qiu {[}49{]} separate writing, storage, routing
and finite-measurement readout in quantum reservoirs using delay-space
quantum Fisher information. These connect resource allocation and memory
geometry in different observation models.

\subsection{Conclusion}\label{conclusion}

We characterise how noisy linear recurrent systems allocate information
during writing, transfer it to a finite store, and preserve it after
writing ends. The finite-horizon Fisher budget constrains non-normality
to redistributing directional information. Coupling and closure
covariance determine which direction is useful at the store. After
writing, an isolated store under an invertible hold preserves exact
Fisher information. An implemented decoder can still lose accuracy
unless it accounts for the changed coordinates. Behavioural optimization
of input masks and linear readouts on fixed linear-Gaussian carriers
approached the task-specific analytic optimum. A separate block of
previously unused carriers reproduced the objective-specific separation.
The development exposure of the earlier studies and the execution
history of each study are disclosed. Learning the recurrent dynamics and
extending these results to nonlinear systems remain open. The analysis
supplies separate criteria for write-direction selection, store
admission, and post-write retention and decoder access.

\subsection{11. Reproducibility}\label{reproducibility}

The release provides the commands below. A successful hash check
verifies the package; it is not an independent repetition of an
experiment. In every fresh extraction, verify the package before running
any other command:

\begin{Shaded}
\begin{Highlighting}[]
\ExtensionTok{shasum} \AttributeTok{{-}a}\NormalTok{ 256 }\AttributeTok{{-}c}\NormalTok{ SHA256SUMS}
\BuiltInTok{export} \VariableTok{PYTHONDONTWRITEBYTECODE}\OperatorTok{=}\NormalTok{1}
\end{Highlighting}
\end{Shaded}

Use one extraction for the analysis and figure rows. In that extraction
only, create the two input-directory links before the first row:

\begin{Shaded}
\begin{Highlighting}[]
\FunctionTok{ln} \AttributeTok{{-}s}\NormalTok{ nc1\_nc2\_confirmatory results/confirmatory}
\FunctionTok{ln} \AttributeTok{{-}s}\NormalTok{ nc3c\_confirmatory results/nc3c}
\end{Highlighting}
\end{Shaded}

Run each training row in its own separate, disposable extraction, after
the checksum check and without these links. The training commands then
create new result directories rather than writing into the released
data. Run each row's commands in the order shown. The first NC-3C block
is already recorded as exercised, so its historical fresh-launch command
is not offered as a new launch; its released outputs can be reanalysed
in the first row. Replaying the second block is a reproduction, not an
additional outcome-unseen study.

{\def\LTcaptype{none} 
\begin{longtable}[]{@{}
  >{\raggedright\arraybackslash}p{(\linewidth - 2\tabcolsep) * \real{0.5000}}
  >{\raggedright\arraybackslash}p{(\linewidth - 2\tabcolsep) * \real{0.5000}}@{}}
\toprule\noalign{}
\begin{minipage}[b]{\linewidth}\raggedright
purpose
\end{minipage} & \begin{minipage}[b]{\linewidth}\raggedright
command
\end{minipage} \\
\midrule\noalign{}
\endhead
\bottomrule\noalign{}
\endlastfoot
after creating the two links above, recompute the verdicts and
manuscript tables from the released outputs &
\texttt{python\ code/nc\_verdict.py\ -\/-out\ NC1\_VERDICT\_RECOMPUTED.json};
\texttt{python\ code/nc3c\_verdict.py\ -\/-out\ NC3C\_VERDICT\_RECOMPUTED.json};
\texttt{python\ code/make\_claim\_ledger.py} \\
check the Cesàro identification (\emph{Classical ingredients}, before
§3.3) & \texttt{python\ code/check\_cesaro\_identification.py} \\
regenerate all manuscript figures and supplementary plots from the
released CSVs into new directories &
\texttt{python\ code/make\_public\_figures.py\ -\/-out\ reproduced/public\_figures};
\texttt{python\ code/make\_nc\_figures.py\ -\/-out\ reproduced/nc\_figures};
\texttt{python\ code/make\_section7\_figures.py\ -\/-out\ reproduced/section7\_figures} \\
rerun NC-1/NC-2 and exploratory training in a disposable extraction
without input links & \texttt{python\ code/run\_stage.py\ smoke};
\texttt{python\ code/run\_stage.py\ pilot};
\texttt{python\ code/run\_stage.py\ confirmatory};
\texttt{python\ code/run\_exploratory.py\ -\/-amplitude\ 1.5\ -\/-train-horizon\ 128} \\
rerun the second carrier block of §7.4.1 in another disposable
extraction without input links &
\texttt{python\ code/run\_nc3c.py\ -\/-block\ 16-31\ -\/-results-dir\ results/nc3c\_block2\_20260920};
\texttt{python\ code/nc3c\_verdict.py\ -\/-results\ results/nc3c\_block2\_20260920\ -\/-out\ results/nc3c\_block2\_20260920/VERDICT.json} \\
verify that the released files remain unchanged &
\texttt{shasum\ -a\ 256\ -c\ SHA256SUMS} \\
\end{longtable}
}

Preregistration and provenance materials are included under
\texttt{preregistration/} and \texttt{provenance/}:

\begin{itemize}
\tightlist
\item
  the NC-1 protocol and its machine-readable preregistration;
\item
  the NC-3C protocol and its machine-readable preregistration;
\item
  the protocol amendments;
\item
  the preregistration-exposure disclosure.
\end{itemize}

Environment: Python 3.11 or 3.12, NumPy \(\ge\) 2.3, SciPy \(\ge\) 1.17,
Matplotlib \(\ge\) 3.11. PyTorch is optional and used only for the
cross-implementation parity control.

\subsubsection{11.1 What reproduction means for this
build}\label{what-reproduction-means-for-this-build}

Carrier matrices are generated through \texttt{numpy.linalg.qr}, which
is not bit-reproducible across LAPACK implementations. Running the
released code with identical seeds on macOS/Accelerate and on
Linux/OpenBLAS produced carriers agreeing to \texttt{9.5e-15} absolute
and \texttt{lambda\_max} to \texttt{1.8e-15} relative, with different
SHA-256 digests. Reproduction is therefore defined on the portable
numerical invariants recorded in every run record, not on byte equality
of the carrier arrays. Each record also notes the BLAS library of the
machine that produced it.

The released runner may adapt file paths and drop internal dependencies.
It is a separate wrapper with its own hash, which is not the hash of the
original scientific run. The numerical implementations are preserved;
public source metadata and dependency-identity edits are mapped to the
original hashes in MANIFEST.json, and numerical parity is checked.

\subsection{Acknowledgements and AI-assistance
statement}\label{acknowledgements-and-ai-assistance-statement}

AI tools were used, under the author's direction, for derivation
checking, code, literature organization and drafting. The author
conceived and selected the research questions, verified the proof
arguments, re-executed the computations and takes responsibility for the
manuscript.

\subsection{References}\label{references}

\begin{enumerate}
\def\labelenumi{\arabic{enumi}.}
\item
  S. Ganguli, D. Huh, H. Sompolinsky. Memory traces in dynamical
  systems. \emph{Proc. Natl. Acad. Sci. USA} 105(48):18970--18975
  (2008). doi:10.1073/pnas.0804451105.
\item
  G. Hennequin, T. P. Vogels, W. Gerstner. Non-normal amplification in
  random balanced neuronal networks. \emph{Phys. Rev.~E} 86:011909
  (2012). arXiv:1204.2945.
\item
  A. E. Orhan, X. Pitkow. Improved memory in recurrent neural networks
  with sequential non-normal dynamics. \emph{ICLR 2020}.
  arXiv:1905.13715.
\item
  M. Asllani, R. Lambiotte, T. Carletti. Structure and dynamical
  behavior of non-normal networks. \emph{Sci. Adv.} 4(12):eaau9403
  (2018).
\item
  G. Kerg, K. Goyette, M. Puelma Touzel, G. Gidel, E. Vorontsov, Y.
  Bengio, G. Lajoie. Non-normal recurrent neural network (nnRNN):
  learning long time dependencies while improving expressivity with
  transient dynamics. \emph{NeurIPS 32} (2019). arXiv:1905.12080.
\item
  M. S. Goldman. Memory without feedback in a neural network.
  \emph{Neuron} 61(4):621--634 (2009).
\item
  M. MacKay, P. Vicol, J. Ba, R. Grosse. Reversible recurrent neural
  networks. \emph{NeurIPS 31} (2018). arXiv:1810.10999.
\item
  Z. Li, J. Han, W. E, Q. Li. On the curse of memory in recurrent neural
  networks: approximation and optimization analysis. \emph{ICLR 2021}.
  arXiv:2009.07799.
\item
  T. K. Rusch, S. Mishra. UnICORNN: a recurrent model for learning very
  long time dependencies. \emph{ICML 2021}. arXiv:2103.05487.
\item
  T. Haruna, K. Nakajima. Memory uncertainty relation and harmonic
  memory in random recurrent networks. arXiv:2605.24628 (2026).
\item
  Supplementary Material S1. Finite-horizon Fisher-memory scripts,
  result tables, run records and the public reproduction entry point
  released with this paper:
  \url{https://github.com/jeonghoon-ad/finite-horizon-fisher-memory}
  (release v1.0).
\item
  P. Tiňo. Fisher memory of linear Wigner echo state networks.
  \emph{ESANN 2017}, pp.~87--92, i6doc.com, ISBN 978-287587039-1.
\item
  O. L. White, D. D. Lee, H. Sompolinsky. Short-term memory in
  orthogonal neural networks. \emph{Phys. Rev.~Lett.} 92(14):148102
  (2004). arXiv:cond-mat/0402452.
\item
  T. Toyoizumi, L. F. Abbott. Beyond the edge of chaos: amplification
  and temporal integration by recurrent networks in the chaotic regime.
  \emph{Phys. Rev.~E} 84:051908 (2011).
\item
  H. Jaeger. Short term memory in echo state networks. GMD Report 152,
  GMD -- Forschungszentrum Informationstechnik, Sankt Augustin (2002).
\item
  M. Hermans, B. Schrauwen. Memory in linear recurrent neural networks
  in continuous time. \emph{Neural Networks} 23(3):341--355 (2010).
  doi:10.1016/j.neunet.2009.08.008.
\item
  H. Jaeger. The ``echo state'' approach to analysing and training
  recurrent neural networks, with an erratum note. GMD Report 148,
  German National Research Center for Information Technology (2001;
  corrected 2010).
\item
  J. Guan, T. Kubota, Y. Kuniyoshi, K. Nakajima. How noise affects
  memory in linear recurrent networks. \emph{Phys. Rev.~Research}
  7:023049 (2025). arXiv:2409.03187.
\item
  J. Dambre, D. Verstraeten, B. Schrauwen, S. Massar. Information
  processing capacity of dynamical systems. \emph{Sci. Rep.} 2:514
  (2012). doi:10.1038/srep00514.
\item
  F. Wang, P. Benner, J. Heiland. Partial observation of linear systems
  with the Mori-Zwanzig formalism. arXiv:2606.23341 (2026).
\item
  Z. Kang, S. Shirasaka, H. Suzuki. Optimizing input mask for maximum
  memory performance of time-delay reservoir subjected to state noise.
  \emph{Nonlinear Theory and Its Applications, IEICE} 12(4):662
  ff.~(2021). doi:10.1587/nolta.12.662.
\item
  J. Ba, G. Hinton, V. Mnih, J. Z. Leibo, C. Ionescu. Using fast weights
  to attend to the recent past. \emph{NeurIPS 29} (2016).
  arXiv:1610.06258.
\item
  M. Beck, K. Pöppel, M. Spanring, A. Auer, O. Prudnikova, M. Kopp, G.
  Klambauer, J. Brandstetter, S. Hochreiter. xLSTM: Extended long
  short-term memory. \emph{NeurIPS 37} (2024). arXiv:2405.04517.
\item
  M. Dowling, H. Jeon, C. Savin, I. M. Park. Memory by design:
  probabilistic sequence layers. arXiv:2605.31163 (2026).
\item
  P. Becker, H. Pandya, G. Gebhardt, C. Zhao, J. Taylor, G. Neumann.
  Recurrent Kalman networks: factorized inference in high-dimensional
  deep feature spaces. \emph{ICML 2019}. arXiv:1905.07357.
\item
  J. Beuria, A. Shukla. Lindblad-inspired multi-timescale reservoir
  computing with separable rotation and dissipation. arXiv:2608.04028
  (2026).
\item
  A. Behrouz, P. Zhong, V. Mirrokni. Titans: learning to memorize at
  test time. arXiv:2501.00663 (2025).
\item
  Y. Shi, C. R. Rojas. Universal estimation of the Fisher information
  for processes with memory. arXiv:2609.14582 (2026).
\item
  B. Boyacıoğlu, F. van Breugel. Fisher information and stochastic
  observability for state estimation in linear systems. arXiv:2410.19975
  (2024).
\item
  G. P. Gehér. Characterisation of Cesàro and L-asymptotic limits of
  matrices. \emph{Linear and Multilinear Algebra} 63(4):788--805 (2015).
  arXiv:1407.1275. doi:10.1080/03081087.2014.899359.
\item
  G. P. Gehér. Asymptotic behaviour of Hilbert space operators with
  applications. Dissertation, arXiv:1505.07205 (2015). Chapter 3
  restates the results of {[}30{]}.
\item
  G. Baggio, V. Rutten, G. Hennequin and S. Zampieri. Efficient
  communication over complex dynamical networks: The role of matrix
  non-normality. \emph{Science Advances} 6:eaba2282 (2020).
  doi:10.1126/sciadv.aba2282.
\item
  D. C. Hoaglin and R. E. Welsch. The hat matrix in regression and
  ANOVA. \emph{The American Statistician} 32(1):17--22 (1978).
  doi:10.1080/00031305.1978.10479237. Author working-paper version
  (1977): https://hdl.handle.net/1721.1/1920.
\item
  A. G. Kachurovskii. The rate of convergence in ergodic theorems.
  \emph{Russian Mathematical Surveys} 51(4):653--703 (1996).
\item
  M. Aloisio, S. L. de Carvalho, C. R. de Oliveira and E. Souza. On
  spectral measures and convergence rates in von Neumann's Ergodic
  Theorem. arXiv:2209.05290 (2022; version 2, 2023).
\item
  A. J. Short and T. C. Farrelly. Quantum equilibration in finite time.
  \emph{New Journal of Physics} 14:013063 (2012).
  doi:10.1088/1367-2630/14/1/013063.
\item
  J. P. Stroud, K. Watanabe, T. Suzuki, M. G. Stokes and M. Lengyel.
  Optimal information loading into working memory explains dynamic
  coding in the prefrontal cortex. \emph{PNAS} 120(48):e2307991120
  (2023). doi:10.1073/pnas.2307991120.
\item
  J. Zylberberg, A. Pouget, P. E. Latham and E. Shea-Brown. Robust
  information propagation through noisy neural circuits. \emph{PLOS
  Computational Biology} 13(4):e1005497 (2017).
  doi:10.1371/journal.pcbi.1005497.
\item
  X. Luo and I. Hoteit. Covariance inflation in the ensemble Kalman
  filter: a residual nudging perspective and some implications.
  arXiv:1305.4496 (2013).
\item
  R. Moreno-Bote, J. Beck, I. Kanitscheider, X. Pitkow, P. Latham and A.
  Pouget. Information-limiting correlations. \emph{Nature Neuroscience}
  17:1410--1417 (2014). doi:10.1038/nn.3807.
\item
  Y. Burak and I. R. Fiete. Fundamental limits on persistent activity in
  networks of noisy neurons. \emph{PNAS} 109(43):17645--17650 (2012).
  doi:10.1073/pnas.1117386109.
\item
  L. Cui, K. Nakajima and K. Aihara. Optimal Memory Encoding Through
  Fluctuation--Response Structure. arXiv:2603.21666 (2026).
\item
  J.-R. King and S. Dehaene. Characterizing the dynamics of mental
  representations: the temporal generalization method. \emph{Trends in
  Cognitive Sciences} 18(4):203--210 (2014).
  doi:10.1016/j.tics.2014.01.002.
\item
  J. D. Murray et al.~Stable population coding for working memory
  coexists with heterogeneous neural dynamics in prefrontal cortex.
  \emph{PNAS} 114(2):394--399 (2017). doi:10.1073/pnas.1619449114.
\item
  K. Aitken, M. Garrett, S. Olsen and S. Mihalas. The geometry of
  representational drift in natural and artificial neural networks.
  \emph{PLOS Computational Biology} 18(11):e1010716 (2022).
  doi:10.1371/journal.pcbi.1010716.
\item
  A. D. Degenhart et al.~Stabilization of a brain-computer interface via
  the alignment of low-dimensional spaces of neural activity.
  \emph{Nature Biomedical Engineering} 4:672--685 (2020).
  doi:10.1038/s41551-020-0542-9.
\item
  V. Joshi et al.~Accurate deep neural network inference using
  computational phase-change memory. \emph{Nature Communications}
  11:2473 (2020). doi:10.1038/s41467-020-16108-9.
\item
  Q. Bai, Z. Li and J. Kou. Fisher-information retention under local
  driving in non-Hermitian feed-forward chains. arXiv:2609.04996 (2026).
\item
  C. Wang and X. Qiu. Fisher-Orthogonal Memory in Quantum Reservoir
  Computing. arXiv:2607.29219 (2026; version 2).
\end{enumerate}

\subsection{Appendix A. Proofs}\label{appendix-a.-proofs}

\subsubsection{A.1 Trace budget}\label{a.1-trace-budget}

Because \(C_n=\sum_{j<n}W^j(W^j)^\top\succeq I\), \[
\operatorname{tr}M_n
=\sum_{k<n}\operatorname{tr}\!\left(C_n^{-1}W^k(W^k)^\top\right)
=\operatorname{tr}(C_n^{-1}C_n)=N.
\]

\subsubsection{A.2 Normal isotropy}\label{a.2-normal-isotropy}

For normal \(W=U\Lambda U^*\), \[
C_n=U\left(\sum_{j<n}|\Lambda|^{2j}\right)U^*.
\] The diagonal contribution of eigenvalue \(\mu_i\) at lag \(k\) is \[
|\mu_i|^{2k}/\sum_{j<n}|\mu_i|^{2j},
\] whose sum over \(k<n\) is one. Therefore \(M_n=I\).

\subsubsection{A.3 Uniform tail bounds}\label{a.3-uniform-tail-bounds}

Bi-power-boundedness gives \[
nK_-^{-2}I\preceq C_n\preceq nK_+^2I
\] and therefore \[
\frac1{nK_+^2}I\preceq C_n^{-1}\preceq\frac{K_-^2}{n}I.
\] Together with \(K_-^{-1}\le\|W^kv\|\le K_+\), this yields Theorem
3.3a.

\subsubsection{A.4 Cesàro--commutant
limit}\label{a.4-cesuxe0rocommutant-limit}

A finite-dimensional bi-power-bounded real \(W\) is similar to an
orthogonal matrix. One self-contained construction uses \[
H_n=\frac1n\sum_{j<n}(W^j)^\top W^j.
\] Any convergent subsequence has a positive-definite limit \(H\)
satisfying \(W^\top HW=H\); then \(Q=H^{1/2}WH^{-1/2}\) is orthogonal.

With \(W=SQS^{-1}\) and \(A=S^{-1}S^{-\top}\), \[
C_n=nSA_nS^\top,\qquad
A_n=\frac1n\sum_{j<n}Q^jAQ^{-j}.
\] The Cesàro mean kills cross terms between distinct eigenvalue groups
of \(Q\) and converges to \(\bar A=\Pi_{\operatorname{Comm}(Q)}A\).
Substitution gives Theorem 3.3b.

For completeness, let \(B\succ0\) be real with \(\operatorname{tr}B=N\).
There is a real orthogonal \(V\) such that \(D=V^\top BV\) has unit
diagonal: choose a unit vector with Rayleigh quotient one, which exists
between the extreme eigenvalues, and repeat on its orthogonal
complement, whose compression has trace equal to its dimension. This
induction also gives \(B=UU^\top\) with \(U=B^{1/2}V\) having unit
columns, and \(U^\top U=V^\top BV\). Choose \(Q\) as distinct planar
rotations with angles in \((0,\pi)\), plus a single fixed axis when
\(N\) is odd. Each diagonal plane block of \(D\) has trace two, so its
conjugation average is the identity; cross terms between different
rotation spectra vanish. Consequently
\(n^{-1}\sum_{j<n}Q^jD Q^{-j}\to I\). Set \[
W=B^{-1/2}VQV^\top B^{1/2}.
\] This real carrier is similar to an orthogonal matrix and satisfies
\(C_n/n\to B^{-1}\). The identification above therefore gives
\(M_\infty=B\), proving real attainability without a complex-to-real
dimension change.

\subsubsection{A.5 Finite-horizon
certification}\label{a.5-finite-horizon-certification}

\textbf{Empty-gap branch.} When \(Q\) has a single distinct eigenvalue
the minimum defining \(\Delta\) is over an empty set and the gap formula
has an undefined denominator. For a real orthogonal \(Q\) this branch is
\(Q=\pm I\), where the commutant is everything, \(A_n=\bar A=A\)
exactly, and the certified error is zero. That case is handled by this
statement rather than by the bound. Let
\(Q=\sum_\lambda\lambda P_\lambda\). Then \[
A_n-\bar A
=
\sum_{\lambda\ne\mu}
c_n(\lambda\bar\mu)P_\lambda A P_\mu,
\quad
c_n(z)=\frac{1-z^n}{n(1-z)}.
\] Since \(|c_n(z)|\le2/(n|1-z|)\), and the spectral blocks are
orthogonal in Frobenius inner product, \[
\|A_n-\bar A\|_F
\le
\frac{2}{n\Delta}\|A-\bar A\|_F.
\] If \(\|\bar A^{-1}\|\delta_n<1\), the standard inverse-perturbation
bound gives \[
\|A_n^{-1}-\bar A^{-1}\|
\le
\frac{\|\bar A^{-1}\|^2\delta_n}
{1-\|\bar A^{-1}\|\delta_n}.
\] Finally, \[
M_n
=
S^{-\top}
\left(\frac1n\sum_{k<n}Q^{-k}A_n^{-1}Q^k\right)
S^{-1},
\] and averaging orthogonal conjugates does not increase the operator
norm.

\subsubsection{A.6 Two-dimensional
corollary}\label{a.6-two-dimensional-corollary}

For an irreducible planar rotation the symmetric commutant is the scalar
matrices, so \[
\bar A=\tfrac12\operatorname{tr}(A)I.
\] Taking \(S=D^{-1}\), \(A=D^2=\operatorname{diag}(1,c)\), gives the
formula in §3.5.

\subsubsection{A.7 Horizon-two identity}\label{a.7-horizon-two-identity}

\[
C_2=I+WW^\top,
\] and the push-through identity gives \[
M_2
=
I+(I+WW^\top)^{-1}-(I+W^\top W)^{-1}.
\] Thus \(M_2=I\) iff \(W\) is normal. Since \(\operatorname{tr}M_2=N\),
every non-normal \(W\) has at least one direction above one and one
below one at horizon two.

\subsubsection{A.8 Store budget}\label{a.8-store-budget}

Let \(C_{\mathrm{sig}}=\sum_{t\in I}L_tL_t^\top\). Under the hypotheses
of Proposition 5.1, \[
0\preceq C_{\mathrm{sig}}\preceq C_{ss}.
\] Therefore \[
\operatorname{tr}M_{\mathrm{store}}
=
\operatorname{tr}(C_{ss}^{-1}C_{\mathrm{sig}})
\le
\operatorname{tr}I_d=d.
\]

\subsubsection{A.9 Exact and approximate
isolation}\label{a.9-exact-and-approximate-isolation}

Exact invariance follows by direct congruence: \[
(A P)^\top(A\Sigma A^\top)^{-1}(A P)=P^\top\Sigma^{-1}P.
\] For the approximate bound, \[
\Sigma_h\preceq(1+\alpha_h)A_h\Sigma_0A_h^\top
\] implies \[
\Sigma_h^{-1}\succeq\frac1{1+\alpha_h}A_h^{-\top}\Sigma_0^{-1}A_h^{-1},
\] which yields \(J_h\ge J_0/(1+\alpha_h)\). Since \(R_h\succeq0\), also
\(J_h\le J_0\).

For the signal-aligned equality case, put \(B=A_h\Sigma_0A_h^\top\) and
\(p=A_hp_0\), so \(p^\top B^{-1}p=J_0\). Sherman--Morrison gives \[
p^\top(B+\epsilon pp^\top)^{-1}p
=J_0-\frac{\epsilon J_0^2}{1+\epsilon J_0}
=\frac{J_0}{1+\epsilon J_0}.
\] The whitened contamination has largest eigenvalue \(\epsilon J_0\),
so the required covariance bound is attained with
\(\alpha_h=\epsilon J_0\).

\subsubsection{A.10 Scalar GLS readout}\label{a.10-scalar-gls-readout}

For \(x=p s+\eta\), \(\eta\sim\mathcal N(0,\Sigma)\), the unbiased
minimum-variance linear coefficient is \[
a=\frac{\Sigma^{-1}p}{p^\top\Sigma^{-1}p},
\] and its error variance is \(1/(p^\top\Sigma^{-1}p)\). Under an
invertible hold, inverse-adjoint propagation, state unwarping and GLS
recomputation are algebraically identical.

\subsubsection{A.11 Binary Gaussian endpoint
accuracy}\label{a.11-binary-gaussian-endpoint-accuracy}

\textbf{Proposition A.11.} Let \(x\mid y\sim\mathcal N(y\,a\,p,\Sigma)\)
with equal priors on \(y\in\{-1,+1\}\), \(a>0\), and \(\Sigma\succ0\)
independent of \(y\). Put \(J=p^\top\Sigma^{-1}p\). If \(J>0\) the
log-likelihood ratio is \(2a\,p^\top\Sigma^{-1}x\), and writing
\(d=p^\top\Sigma^{-1}x\) we have \(d\mid y\sim\mathcal N(y\,aJ,\,J)\).
The likelihood-ratio rule therefore has accuracy

\[
\Pr(y\,d>0)=\Pr\bigl[\mathcal N(aJ,J)>0\bigr]=\Phi\!\left(a\sqrt J\right).
\]

If \(J=0\) the two class-conditional laws coincide and the optimal
accuracy is \(1/2\).

\emph{Proof.} The log-density difference is
\(-\tfrac12(x-ap)^\top\Sigma^{-1}(x-ap)+\tfrac12(x+ap)^\top\Sigma^{-1}(x+ap)
=2a\,p^\top\Sigma^{-1}x\). Under \(y=+1\), \(d\) is a linear image of
\(x\) with mean \(a\,p^\top\Sigma^{-1}p=aJ\) and variance
\(p^\top\Sigma^{-1}\Sigma\Sigma^{-1}p=J\). With equal priors the
threshold is zero, and
\(\Pr[\mathcal N(aJ,J)>0]=\Phi(aJ/\sqrt J)=\Phi(a\sqrt J)\). The case
\(y=-1\) is symmetric. If \(J=0\) then \(\Sigma^{-1}p=0\), the two laws
are identical and no rule beats \(1/2\). \(\square\)

The quantity \(J\) is the Fisher information for the associated
continuous location parameter; the proposition relates it to the
specified binary task. It does not assert that an arbitrary decoder
attains Bayes accuracy, which is exactly the gap §7.6 measures.

\subsection{Appendix B. Replicated directional census, full statistics
(8 independent matrix instances per configuration; generated from the
result
CSVs)}\label{appendix-b.-replicated-directional-census-full-statistics-8-independent-matrix-instances-per-configuration-generated-from-the-result-csvs}

B.1 Spectral and gain statistics at \(n=2048\) (\(\lambda_{max}\),
\(\lambda_{min}\) of \(M_{2048}\); \(\mathrm{tr}M_{2048}/N\);
finite-window \(\sigma_{4096}=\max_{j\le4096}\|W^{\pm j}\|_2\) reported
as \texttt{K\_plus}, \texttt{K\_minus}).

{\def\LTcaptype{none} 
\begin{longtable}[]{@{}
  >{\raggedright\arraybackslash}p{(\linewidth - 14\tabcolsep) * \real{0.1250}}
  >{\raggedright\arraybackslash}p{(\linewidth - 14\tabcolsep) * \real{0.1250}}
  >{\raggedright\arraybackslash}p{(\linewidth - 14\tabcolsep) * \real{0.1250}}
  >{\raggedright\arraybackslash}p{(\linewidth - 14\tabcolsep) * \real{0.1250}}
  >{\raggedright\arraybackslash}p{(\linewidth - 14\tabcolsep) * \real{0.1250}}
  >{\raggedright\arraybackslash}p{(\linewidth - 14\tabcolsep) * \real{0.1250}}
  >{\raggedright\arraybackslash}p{(\linewidth - 14\tabcolsep) * \real{0.1250}}
  >{\raggedright\arraybackslash}p{(\linewidth - 14\tabcolsep) * \real{0.1250}}@{}}
\toprule\noalign{}
\begin{minipage}[b]{\linewidth}\raggedright
configuration
\end{minipage} & \begin{minipage}[b]{\linewidth}\raggedright
metric
\end{minipage} & \begin{minipage}[b]{\linewidth}\raggedright
median
\end{minipage} & \begin{minipage}[b]{\linewidth}\raggedright
Q1
\end{minipage} & \begin{minipage}[b]{\linewidth}\raggedright
Q3
\end{minipage} & \begin{minipage}[b]{\linewidth}\raggedright
min
\end{minipage} & \begin{minipage}[b]{\linewidth}\raggedright
max
\end{minipage} & \begin{minipage}[b]{\linewidth}\raggedright
n
\end{minipage} \\
\midrule\noalign{}
\endhead
\bottomrule\noalign{}
\endlastfoot
A0 & lambda\_max & 1 & 1 & 1 & 1 & 1 & 8 \\
A0 & lambda\_min & 1 & 1 & 1 & 1 & 1 & 8 \\
A0 & trace\_over\_N & 1 & 1 & 1 & 1 & 1 & 8 \\
A0 & K\_plus & 1 & 1 & 1 & 1 & 1 & 8 \\
A0 & K\_minus & 1 & 1 & 1 & 1 & 1 & 8 \\
SQS-c2 & lambda\_max & 1.83134 & 1.78732 & 1.83805 & 1.77414 & 1.85209 &
8 \\
SQS-c2 & lambda\_min & 0.470081 & 0.460785 & 0.473855 & 0.457123 &
0.475268 & 8 \\
SQS-c2 & trace\_over\_N & 1 & 1 & 1 & 1 & 1 & 8 \\
SQS-c2 & K\_plus & 1.81775 & 1.8147 & 1.82557 & 1.80592 & 1.83847 & 8 \\
SQS-c2 & K\_minus & 1.82148 & 1.81407 & 1.82802 & 1.79142 & 1.83944 &
8 \\
SQS-c5 & lambda\_max & 3.13892 & 3.08368 & 3.23637 & 3.03958 & 3.29784 &
8 \\
SQS-c5 & lambda\_min & 0.13293 & 0.132166 & 0.135387 & 0.125704 &
0.142916 & 8 \\
SQS-c5 & trace\_over\_N & 1 & 1 & 1 & 1 & 1 & 8 \\
SQS-c5 & K\_plus & 4.14197 & 4.11157 & 4.18571 & 3.94781 & 4.27934 &
8 \\
SQS-c5 & K\_minus & 4.13054 & 4.09408 & 4.21292 & 4.01679 & 4.30329 &
8 \\
SQS-c10 & lambda\_max & 4.31172 & 4.23504 & 4.46172 & 4.09835 & 4.63897
& 8 \\
SQS-c10 & lambda\_min & 0.0451568 & 0.0436136 & 0.0462884 & 0.0416706 &
0.0483073 & 8 \\
SQS-c10 & trace\_over\_N & 1 & 1 & 1 & 1 & 1 & 8 \\
SQS-c10 & K\_plus & 7.87978 & 7.79845 & 8.12216 & 7.67695 & 8.41416 &
8 \\
SQS-c10 & K\_minus & 8.03709 & 7.71964 & 8.12114 & 7.60136 & 8.22991 &
8 \\
SQS-c20 & lambda\_max & 5.4914 & 5.3822 & 5.54088 & 5.24643 & 5.63677 &
8 \\
SQS-c20 & lambda\_min & 0.0140049 & 0.0137212 & 0.0143716 & 0.0128531 &
0.0149251 & 8 \\
SQS-c20 & trace\_over\_N & 1 & 1 & 1 & 1 & 1 & 8 \\
SQS-c20 & K\_plus & 15.0304 & 14.8175 & 15.4953 & 14.6143 & 15.7217 &
8 \\
SQS-c20 & K\_minus & 15.6207 & 15.4536 & 15.8108 & 14.8456 & 16.7114 &
8 \\
SQS-c50 & lambda\_max & 6.56331 & 6.46044 & 6.72827 & 6.34304 & 6.88027
& 8 \\
SQS-c50 & lambda\_min & 0.00292857 & 0.00271839 & 0.00307291 &
0.00257505 & 0.00342859 & 8 \\
SQS-c50 & trace\_over\_N & 1 & 1 & 1 & 1 & 1 & 8 \\
SQS-c50 & K\_plus & 35.0751 & 33.9516 & 35.7027 & 31.4454 & 36.4715 &
8 \\
SQS-c50 & K\_minus & 36.0658 & 34.1093 & 36.8579 & 32.9037 & 38.5231 &
8 \\
SQS-c100 & lambda\_max & 7.7197 & 7.56441 & 7.86664 & 7.36578 & 8.06077
& 8 \\
SQS-c100 & lambda\_min & 0.000846588 & 0.000808496 & 0.000871845 &
0.000727769 & 0.000925082 & 8 \\
SQS-c100 & trace\_over\_N & 1 & 1 & 1 & 1 & 1 & 8 \\
SQS-c100 & K\_plus & 68.8515 & 67.5224 & 69.7734 & 65.184 & 70.6132 &
8 \\
SQS-c100 & K\_minus & 67.8479 & 66.3895 & 70.0367 & 65.3716 & 73.1 &
8 \\
SQS-c35-heldout & lambda\_max & 6.10729 & 6.00661 & 6.2536 & 5.89206 &
6.3313 & 8 \\
SQS-c35-heldout & lambda\_min & 0.00539445 & 0.00510687 & 0.00556909 &
0.00497085 & 0.0058889 & 8 \\
SQS-c35-heldout & trace\_over\_N & 1 & 1 & 1 & 1 & 1 & 8 \\
SQS-c35-heldout & K\_plus & 25.5792 & 24.9064 & 26.1824 & 24.2024 &
28.1703 & 8 \\
SQS-c35-heldout & K\_minus & 25.865 & 25.2868 & 26.8344 & 24.0368 &
28.2466 & 8 \\
ELL-c4-theta-0.7 & lambda\_max & 1.6 & 1.6 & 1.6 & 1.6 & 1.6 & 8 \\
ELL-c4-theta-0.7 & lambda\_min & 0.4 & 0.4 & 0.4 & 0.4 & 0.4 & 8 \\
ELL-c4-theta-0.7 & trace\_over\_N & 1 & 1 & 1 & 1 & 1 & 8 \\
ELL-c4-theta-0.7 & K\_plus & 2 & 2 & 2 & 2 & 2 & 8 \\
ELL-c4-theta-0.7 & K\_minus & 2 & 2 & 2 & 2 & 2 & 8 \\
ELL-c4-theta-2.0 & lambda\_max & 1.6 & 1.6 & 1.6 & 1.6 & 1.6 & 8 \\
ELL-c4-theta-2.0 & lambda\_min & 0.4 & 0.4 & 0.4 & 0.4 & 0.4 & 8 \\
ELL-c4-theta-2.0 & trace\_over\_N & 1 & 1 & 1 & 1 & 1 & 8 \\
ELL-c4-theta-2.0 & K\_plus & 1.99999 & 1.99999 & 1.99999 & 1.99999 &
1.99999 & 8 \\
ELL-c4-theta-2.0 & K\_minus & 1.99999 & 1.99999 & 1.99999 & 1.99999 &
1.99999 & 8 \\
ELL-c4-theta-sqrt\_c & lambda\_max & 1.6 & 1.6 & 1.6 & 1.6 & 1.6 & 8 \\
ELL-c4-theta-sqrt\_c & lambda\_min & 0.4 & 0.4 & 0.4 & 0.4 & 0.4 & 8 \\
ELL-c4-theta-sqrt\_c & trace\_over\_N & 1 & 1 & 1 & 1 & 1 & 8 \\
ELL-c4-theta-sqrt\_c & K\_plus & 1.99999 & 1.99999 & 1.99999 & 1.99999 &
1.99999 & 8 \\
ELL-c4-theta-sqrt\_c & K\_minus & 1.99999 & 1.99999 & 1.99999 & 1.99999
& 1.99999 & 8 \\
ELL-c100-theta-0.7 & lambda\_max & 1.9802 & 1.9802 & 1.9802 & 1.9802 &
1.9802 & 8 \\
ELL-c100-theta-0.7 & lambda\_min & 0.019802 & 0.019802 & 0.019802 &
0.019802 & 0.019802 & 8 \\
ELL-c100-theta-0.7 & trace\_over\_N & 1 & 1 & 1 & 1 & 1 & 8 \\
ELL-c100-theta-0.7 & K\_plus & 10 & 10 & 10 & 10 & 10 & 8 \\
ELL-c100-theta-0.7 & K\_minus & 10 & 10 & 10 & 10 & 10 & 8 \\
ELL-c100-theta-2.0 & lambda\_max & 1.9802 & 1.9802 & 1.9802 & 1.9802 &
1.9802 & 8 \\
ELL-c100-theta-2.0 & lambda\_min & 0.019802 & 0.019802 & 0.019802 &
0.019802 & 0.019802 & 8 \\
ELL-c100-theta-2.0 & trace\_over\_N & 1 & 1 & 1 & 1 & 1 & 8 \\
ELL-c100-theta-2.0 & K\_plus & 9.99993 & 9.99993 & 9.99993 & 9.99993 &
9.99993 & 8 \\
ELL-c100-theta-2.0 & K\_minus & 9.99993 & 9.99993 & 9.99993 & 9.99993 &
9.99993 & 8 \\
ELL-c100-theta-sqrt\_c & lambda\_max & 1.9802 & 1.9802 & 1.9802 & 1.9802
& 1.9802 & 8 \\
ELL-c100-theta-sqrt\_c & lambda\_min & 0.019802 & 0.019802 & 0.019802 &
0.019802 & 0.019802 & 8 \\
ELL-c100-theta-sqrt\_c & trace\_over\_N & 1 & 1 & 1 & 1 & 1 & 8 \\
ELL-c100-theta-sqrt\_c & K\_plus & 9.99829 & 9.99829 & 9.99829 & 9.99829
& 9.99829 & 8 \\
ELL-c100-theta-sqrt\_c & K\_minus & 9.99829 & 9.99829 & 9.99829 &
9.99829 & 9.99829 & 8 \\
H8-c4 & lambda\_max & 1.59544 & 1.55396 & 1.66172 & 1.46834 & 1.714 &
8 \\
H8-c4 & lambda\_min & 0.548615 & 0.506983 & 0.585517 & 0.472261 &
0.639818 & 8 \\
H8-c4 & trace\_over\_N & 1 & 1 & 1 & 1 & 1 & 8 \\
H8-c4 & K\_plus & 1.69905 & 1.60287 & 1.79612 & 1.51029 & 1.8567 & 8 \\
H8-c4 & K\_minus & 1.69905 & 1.60287 & 1.79612 & 1.51029 & 1.8567 & 8 \\
H8-c100 & lambda\_max & 2.77146 & 2.4617 & 2.92751 & 2.14714 & 3.35084 &
8 \\
H8-c100 & lambda\_min & 0.113819 & 0.10092 & 0.134624 & 0.0681658 &
0.250028 & 8 \\
H8-c100 & trace\_over\_N & 1 & 1 & 1 & 1 & 1 & 8 \\
H8-c100 & K\_plus & 4.93585 & 4.25196 & 5.18659 & 2.89813 & 6.79367 &
8 \\
H8-c100 & K\_minus & 4.93585 & 4.25196 & 5.18659 & 2.89813 & 6.79367 &
8 \\
ELL-c4-theta-1.1 & lambda\_max & 1.6 & 1.6 & 1.6 & 1.6 & 1.6 & 8 \\
ELL-c4-theta-1.1 & lambda\_min & 0.4 & 0.4 & 0.4 & 0.4 & 0.4 & 8 \\
ELL-c4-theta-1.1 & trace\_over\_N & 1 & 1 & 1 & 1 & 1 & 8 \\
ELL-c4-theta-1.1 & K\_plus & 2 & 2 & 2 & 2 & 2 & 8 \\
ELL-c4-theta-1.1 & K\_minus & 2 & 2 & 2 & 2 & 2 & 8 \\
\end{longtable}
}

B.2 Oldest-lag Fisher information in the oracle direction, \(nJ_n(n-1)\)
by horizon.

{\def\LTcaptype{none} 
\begin{longtable}[]{@{}
  >{\raggedright\arraybackslash}p{(\linewidth - 10\tabcolsep) * \real{0.1667}}
  >{\raggedright\arraybackslash}p{(\linewidth - 10\tabcolsep) * \real{0.1667}}
  >{\raggedright\arraybackslash}p{(\linewidth - 10\tabcolsep) * \real{0.1667}}
  >{\raggedright\arraybackslash}p{(\linewidth - 10\tabcolsep) * \real{0.1667}}
  >{\raggedright\arraybackslash}p{(\linewidth - 10\tabcolsep) * \real{0.1667}}
  >{\raggedright\arraybackslash}p{(\linewidth - 10\tabcolsep) * \real{0.1667}}@{}}
\toprule\noalign{}
\begin{minipage}[b]{\linewidth}\raggedright
configuration
\end{minipage} & \begin{minipage}[b]{\linewidth}\raggedright
n=32 med {[}min,max{]}
\end{minipage} & \begin{minipage}[b]{\linewidth}\raggedright
n=128 med {[}min,max{]}
\end{minipage} & \begin{minipage}[b]{\linewidth}\raggedright
n=512 med {[}min,max{]}
\end{minipage} & \begin{minipage}[b]{\linewidth}\raggedright
n=2048 med {[}min,max{]}
\end{minipage} & \begin{minipage}[b]{\linewidth}\raggedright
n=4096 med {[}min,max{]}
\end{minipage} \\
\midrule\noalign{}
\endhead
\bottomrule\noalign{}
\endlastfoot
A0 & 1.0000 {[}1.0000, 1.0000{]} & 1.0000 {[}1.0000, 1.0000{]} & 1.0000
{[}1.0000, 1.0000{]} & 1.0000 {[}1.0000, 1.0000{]} & 1.0000 {[}1.0000,
1.0000{]} \\
SQS-c2 & 1.8087 {[}1.7671, 1.8399{]} & 1.8216 {[}1.7681, 1.8480{]} &
1.8298 {[}1.7728, 1.8505{]} & 1.8309 {[}1.7740, 1.8517{]} & 1.8311
{[}1.7740, 1.8519{]} \\
SQS-c5 & 3.0782 {[}2.9180, 3.1654{]} & 3.1047 {[}3.0034, 3.2607{]} &
3.1297 {[}3.0303, 3.2881{]} & 3.1379 {[}3.0377, 3.2961{]} & 3.1379
{[}3.0385, 3.2972{]} \\
SQS-c10 & 4.1283 {[}3.8057, 4.4853{]} & 4.2511 {[}3.9791, 4.5919{]} &
4.2641 {[}4.1207, 4.6274{]} & 4.3168 {[}4.0891, 4.6366{]} & 4.3102
{[}4.0976, 4.6372{]} \\
SQS-c20 & 5.0889 {[}4.9661, 5.4788{]} & 5.4363 {[}5.1671, 5.6192{]} &
5.4808 {[}5.2417, 5.6027{]} & 5.4862 {[}5.2396, 5.6330{]} & 5.4883
{[}5.2425, 5.6335{]} \\
SQS-c50 & 6.1324 {[}5.9256, 6.5722{]} & 6.4079 {[}6.2504, 6.8209{]} &
6.5396 {[}6.3195, 6.8635{]} & 6.5519 {[}6.3343, 6.8720{]} & 6.5594
{[}6.3398, 6.8768{]} \\
SQS-c100 & 7.4176 {[}6.8130, 7.7072{]} & 7.5664 {[}7.1871, 7.9315{]} &
7.6758 {[}7.3394, 8.0098{]} & 7.7119 {[}7.3729, 8.0450{]} & 7.7130
{[}7.3690, 8.0578{]} \\
SQS-c35-heldout & 5.7492 {[}5.4506, 6.0658{]} & 6.0579 {[}5.8213,
6.3185{]} & 6.0905 {[}5.8614, 6.3021{]} & 6.1007 {[}5.8829, 6.3287{]} &
6.1051 {[}5.8894, 6.3299{]} \\
ELL-c4-theta-0.7 & 1.5825 {[}1.5825, 1.5825{]} & 1.5932 {[}1.5932,
1.5932{]} & 1.5993 {[}1.5993, 1.5993{]} & 1.5994 {[}1.5994, 1.5994{]} &
1.5999 {[}1.5999, 1.5999{]} \\
ELL-c4-theta-2.0 & 1.5801 {[}1.5801, 1.5801{]} & 1.5927 {[}1.5927,
1.5927{]} & 1.5998 {[}1.5998, 1.5998{]} & 1.5997 {[}1.5997, 1.5997{]} &
1.5998 {[}1.5998, 1.5998{]} \\
ELL-c4-theta-sqrt\_c & 1.5801 {[}1.5801, 1.5801{]} & 1.5927 {[}1.5927,
1.5927{]} & 1.5998 {[}1.5998, 1.5998{]} & 1.5997 {[}1.5997, 1.5997{]} &
1.5998 {[}1.5998, 1.5998{]} \\
ELL-c100-theta-0.7 & 1.9450 {[}1.9450, 1.9450{]} & 1.9665 {[}1.9665,
1.9665{]} & 1.9788 {[}1.9788, 1.9788{]} & 1.9790 {[}1.9790, 1.9790{]} &
1.9801 {[}1.9801, 1.9801{]} \\
ELL-c100-theta-2.0 & 1.9407 {[}1.9407, 1.9407{]} & 1.9655 {[}1.9655,
1.9655{]} & 1.9798 {[}1.9798, 1.9798{]} & 1.9797 {[}1.9797, 1.9797{]} &
1.9797 {[}1.9797, 1.9797{]} \\
ELL-c100-theta-sqrt\_c & 2.0064 {[}2.0064, 2.0064{]} & 1.9615 {[}1.9615,
1.9615{]} & 1.9812 {[}1.9812, 1.9812{]} & 1.9803 {[}1.9803, 1.9803{]} &
1.9803 {[}1.9803, 1.9803{]} \\
H8-c4 & 1.5764 {[}1.4490, 1.6915{]} & 1.5953 {[}1.4637, 1.7099{]} &
1.6069 {[}1.4676, 1.7130{]} & 1.5951 {[}1.4681, 1.7138{]} & 1.5954
{[}1.4682, 1.7139{]} \\
H8-c100 & 2.5878 {[}1.9361, 3.5008{]} & 2.7431 {[}2.1616, 3.2840{]} &
2.7663 {[}2.1384, 3.3507{]} & 2.7692 {[}2.1512, 3.3501{]} & 2.7709
{[}2.1484, 3.3497{]} \\
ELL-c4-theta-1.1 & 1.5822 {[}1.5822, 1.5822{]} & 1.5996 {[}1.5996,
1.5996{]} & 1.5985 {[}1.5985, 1.5985{]} & 1.5999 {[}1.5999, 1.5999{]} &
1.5999 {[}1.5999, 1.5999{]} \\
\end{longtable}
}

\subsection{Appendix C. Write-path × storage
factorial}\label{appendix-c.-write-path-storage-factorial}

\emph{Readings are given here as short labels; the exact
machine-readable label for each is in the released data
(\texttt{results/.../PREREGISTERED\_READINGS.json}).}

\subsubsection{C.1 Isolation Factorial v2 (the §5 experiment):
pre-registered readings and raw
medians}\label{c.1-isolation-factorial-v2-the-5-experiment-pre-registered-readings-and-raw-medians}

Run record: valid measurement, 296 controls, 0 failed, attempt 1,
elapsed 7.8 s; script SHA \texttt{69acea303c4c14e1…}. Readings (all on
the oracle sub-arm, thresholds fixed before the run):

{\def\LTcaptype{none} 
\begin{longtable}[]{@{}
  >{\raggedright\arraybackslash}p{(\linewidth - 6\tabcolsep) * \real{0.2500}}
  >{\raggedright\arraybackslash}p{(\linewidth - 6\tabcolsep) * \real{0.2500}}
  >{\raggedright\arraybackslash}p{(\linewidth - 6\tabcolsep) * \real{0.2500}}
  >{\raggedright\arraybackslash}p{(\linewidth - 6\tabcolsep) * \real{0.2500}}@{}}
\toprule\noalign{}
\begin{minipage}[b]{\linewidth}\raggedright
reading
\end{minipage} & \begin{minipage}[b]{\linewidth}\raggedright
threshold
\end{minipage} & \begin{minipage}[b]{\linewidth}\raggedright
value
\end{minipage} & \begin{minipage}[b]{\linewidth}\raggedright
label
\end{minipage} \\
\midrule\noalign{}
\endhead
\bottomrule\noalign{}
\endlastfoot
paired ratio, conditioned/normal, isolated, store-only \(J_{tot}\),
\(n=4096\) & \textgreater{} 1.05 & median 5.429 {[}3.926, 11.800{]} &
store concentration above one: yes \\
normal isolated, oldest-lag relative spread over horizons & \textless{}
1e-6 & 1.6e-15; medians 0.0337536, 0.0337536, 0.0337536, 0.0337536 &
oldest-lag store info constant: yes \\
normal open, oldest-lag \(n{=}64\) / \(n{=}4096\) & \textgreater= 10 &
118.8 & open-cell oldest-lag decays: yes \\
normal isolated, reach (inputs with \(J^{(s)}>10^{-6}\)) & report & 23
of 24 in every draw & REACH\_COUNT \\
nonnormal isolated, oldest-lag relative spread over horizons &
\textless{} 1e-6 & 1.3e-15; medians 0.1882299, 0.1882299, 0.1882299,
0.1882299 & oldest-lag store info constant: yes \\
nonnormal open, oldest-lag \(n{=}64\) / \(n{=}4096\) & \textgreater= 10
& 88.0 & open-cell oldest-lag decays: yes \\
nonnormal isolated, reach (inputs with \(J^{(s)}>10^{-6}\)) & report &
23 of 24 in every draw & REACH\_COUNT \\
\end{longtable}
}

Raw medians by cell, direction and horizon (store-only total /
store-only oldest \(t{=}0\) / full-state total):

{\def\LTcaptype{none} 
\begin{longtable}[]{@{}
  >{\raggedright\arraybackslash}p{(\linewidth - 12\tabcolsep) * \real{0.1429}}
  >{\raggedright\arraybackslash}p{(\linewidth - 12\tabcolsep) * \real{0.1429}}
  >{\raggedright\arraybackslash}p{(\linewidth - 12\tabcolsep) * \real{0.1429}}
  >{\raggedright\arraybackslash}p{(\linewidth - 12\tabcolsep) * \real{0.1429}}
  >{\raggedright\arraybackslash}p{(\linewidth - 12\tabcolsep) * \real{0.1429}}
  >{\raggedright\arraybackslash}p{(\linewidth - 12\tabcolsep) * \real{0.1429}}
  >{\raggedright\arraybackslash}p{(\linewidth - 12\tabcolsep) * \real{0.1429}}@{}}
\toprule\noalign{}
\begin{minipage}[b]{\linewidth}\raggedright
write
\end{minipage} & \begin{minipage}[b]{\linewidth}\raggedright
storage
\end{minipage} & \begin{minipage}[b]{\linewidth}\raggedright
direction
\end{minipage} & \begin{minipage}[b]{\linewidth}\raggedright
\(n=64\)
\end{minipage} & \begin{minipage}[b]{\linewidth}\raggedright
\(n=256\)
\end{minipage} & \begin{minipage}[b]{\linewidth}\raggedright
\(n=1024\)
\end{minipage} & \begin{minipage}[b]{\linewidth}\raggedright
\(n=4096\)
\end{minipage} \\
\midrule\noalign{}
\endhead
\bottomrule\noalign{}
\endlastfoot
normal & open & oracle & 0.413 / 0.00994 / 1.385 & 0.397 / 0.00199 /
1.354 & 0.405 / 0.00044 / 1.374 & 0.405 / 0.00008 / 1.376 \\
normal & open & random & 0.626 / 0.01550 / 1.719 & 0.630 / 0.00259 /
1.708 & 0.622 / 0.00057 / 1.711 & 0.621 / 0.00016 / 1.711 \\
normal & isolated & oracle & 0.421 / 0.03375 / 1.417 & 0.421 / 0.03375 /
1.420 & 0.421 / 0.03375 / 1.421 & 0.421 / 0.03375 / 1.421 \\
normal & isolated & random & 0.597 / 0.04447 / 1.597 & 0.597 / 0.04447 /
1.597 & 0.597 / 0.04447 / 1.597 & 0.597 / 0.04447 / 1.597 \\
nonnormal & open & oracle & 2.263 / 0.04914 / 6.578 & 2.268 / 0.01055 /
6.770 & 2.287 / 0.00170 / 6.800 & 2.290 / 0.00056 / 6.800 \\
nonnormal & open & random & 0.456 / 0.01244 / 1.449 & 0.471 / 0.00204 /
1.409 & 0.469 / 0.00043 / 1.407 & 0.469 / 0.00007 / 1.407 \\
nonnormal & isolated & oracle & 2.509 / 0.18823 / 6.746 & 2.509 /
0.18823 / 6.734 & 2.509 / 0.18823 / 6.747 & 2.509 / 0.18823 / 6.750 \\
nonnormal & isolated & random & 0.439 / 0.03302 / 1.437 & 0.439 /
0.03302 / 1.443 & 0.439 / 0.03302 / 1.444 & 0.439 / 0.03302 / 1.444 \\
\end{longtable}
}

Invariance controls (draw 0, conditioned carrier; maximum absolute curve
difference from the isometric isolated cell): \(0.9U\) after closure
only: store-only \(5\times10^{-16}\) at \(n\le1024\),
\(1.9\times10^{-1}\) at \(n=4096\) (underflow); full-state
\(5\times10^{-15}\) at \(n=64\), \(\sim0.19\) at \(n\ge256\)
(pseudoinverse threshold). \(I\) after closure: \(\le1.7\times10^{-15}\)
in both readouts at every horizon. \(0.9U\) during the write window:
\(8\times10^{-2}\) store-only at every horizon. Gated controls:
store-only at \(n\in\{64,256,1024\}\) and full-state at \(n=64\), all
passed at \(\le2.7\times10^{-14}\).

\subsubsection{C.2 Numerical checks for §§3.4, 5.3, 6.2 and
6.3}\label{c.2-numerical-checks-for-3.4-5.3-6.2-and-6.3}

Run: \texttt{preprint\_v1\_2\_strengthening\_20260917.py}; script
SHA-256
\texttt{c1fd328db9cea8677f74611d96d1cce9a09b673a096da7098aaba249ab7fa7bc}.

\textbf{Finite-horizon certification, \(c=10\), \(n=2048\), eight paired
instances.}

\begin{itemize}
\tightlist
\item
  relative Frobenius error median: 8.444e-06
\item
  range: 3.558e-06 to 1.792e-05
\item
  every actual operator-norm error below the reported sufficient bound:
  yes
\end{itemize}

\textbf{Store-operator selection.}

\begin{itemize}
\tightlist
\item
  store trace: 16.0
\item
  store-oracle / write-oracle total ratio: median 1.224, range
  1.112--1.582
\item
  oldest-input ratio: median 1.257, range 0.922--1.762
\item
  individual oldest-input losses: 1 of 8
\item
  all selected store directions retain write-block allocation at least
  1.05: yes
\end{itemize}

\textbf{Sampled decoder, 5,000 trials per draw.}

\begin{itemize}
\tightlist
\item
  sampled/theoretical error-variance ratio: median 1.017, range
  0.969--1.032
\item
  maximum discrepancy among exact compensated readings: 7.994e-15
\end{itemize}

\textbf{Approximate isolation.}

\begin{itemize}
\tightlist
\item
  all randomized and aligned bounds pass: yes
\item
  minimum numerical margin: -4.441e-16
\item
  maximum equality-case error: 4.441e-16
\end{itemize}

\subsection{Appendix D. Reproducibility and
artifacts}\label{appendix-d.-reproducibility-and-artifacts}

The public release is available at
\url{https://github.com/jeonghoon-ad/finite-horizon-fisher-memory}
(release v1.0) and contains:

\begin{itemize}
\tightlist
\item
  \texttt{manuscript/}: this manuscript as Markdown, LaTeX and PDF, with
  Figures 1 to 7 in PNG and PDF;
\item
  \texttt{claim\_ledger/}: every reported number with its source
  artifact, its subset and its aggregation named;
\item
  \texttt{provenance/}: the preregistration-exposure disclosure, the
  clean-build verification, the prior-art disposition, the changelogs
  and the source-derivative records;
\item
  \texttt{preregistration/}: the NC-1 and NC-3C protocols, their
  machine-readable preregistrations, and the protocol amendments;
\item
  \texttt{code/}: the scientific modules, the figure scripts, the
  manuscript build script and its LaTeX header, and records of the code
  versions used;
\item
  \texttt{results/}: the confirmatory, pilot, smoke, exploratory and
  parity outputs; the second carrier block of §7.4.1 with its freshness
  check, its seed list and the record of its invalid first execution;
  and the figure inputs;
\item
  \texttt{vendor/}: the v1.2 upstream release this work imports from,
  with public-package metadata and attribution edits recorded in
  MANIFEST.json; numerical data and scientific implementations are
  unchanged;
\item
  \texttt{README.md}, \texttt{LICENSE\_SCOPE.md}, \texttt{SHA256SUMS},
  \texttt{MANIFEST.json}, \texttt{REPRODUCE.md} and
  \texttt{requirements.txt}.
\end{itemize}

The run record for the checks of Appendix C.2 is included in
Supplementary Material S1; \texttt{REPRODUCE.md} gives its location.
Earlier drafts and internal working records are not included.

\end{document}